Published in: Expert Systems With Applications, 2025, Elsevier

Cite as:

BibTeX:

```
@article{SILVA2025128529,
    author   =  {Renato M. Silva and Gregório F. Azevedo and
                 Matheus V. V. Berto and Jean R. Rocha and
                 Eduardo C. Fidelis and Matheus V. Nogueira and
                 Pedro H. Lisboa and Tiago A. Almeida},
    title    =  {Vulnerable road user detection and safety enhancement: A comprehensive survey},
    pages    =  {128529},
    journal  =  {Expert Systems with Applications},
    year     =  {2025},
    issn     =  {0957-4174},
    DOI      =  {10.1016/j.eswa.2025.128529},
}
```

# Vulnerable Road User Detection and Safety Enhancement: A Comprehensive Survey

RENATO M. SILVA, Institute of Mathematics and Computer Science, University of São Paulo – USP, Brazil
GREGÓRIO F. AZEVEDO, Department of Computer Science, Federal University of São Carlos – UFSCar, Brazil
MATHEUS V. V. BERTO, Department of Computer Science, Federal University of São Carlos – UFSCar, Brazil
JEAN R. ROCHA, Department of Computer Science, Federal University of São Carlos – UFSCar, Brazil
EDUARDO C. FIDELIS, Department of Computer Engineering, Facens University, Brazil
MATHEUS V. NOGUEIRA, Department of Computer Engineering, Facens University, Brazil
PEDRO H. LISBOA, Department of Computer Engineering, Facens University, Brazil
TIAGO A. ALMEIDA, Department of Computer Science, Federal University of São Carlos – UFSCar, Brazil

Traffic incidents involving vulnerable road users (VRUs) constitute a significant proportion of global road accidents. Advances in traffic communication ecosystems, coupled with sophisticated signal processing and machine learning techniques, have facilitated the utilization of data from diverse sensors. Despite these advancements and the availability of extensive datasets, substantial progress is required to mitigate traffic casualties. This paper provides a comprehensive survey of state-of-the-art technologies and methodologies to enhance the safety of VRUs. The study investigates the communication networks between vehicles and VRUs, emphasizing the integration of advanced sensors and the availability of relevant datasets. It explores preprocessing techniques and data fusion methods to enhance sensor data quality. Furthermore, our study assesses critical simulation environments essential for developing and testing VRU safety systems. Our research also highlights recent advances in VRU detection and classification algorithms, addressing challenges such as variable environmental conditions. Additionally, we cover cutting-edge research in predicting VRU intentions and behaviors, which is mandatory for proactive collision avoidance strategies. Through this survey, we aim to provide a comprehensive understanding of the current landscape of VRU safety technologies, identifying areas of progress and areas needing further research and development.

Authors' addresses: Renato M. Silva, renatoms@icmc.usp.br, Institute of Mathematics and Computer Science, University of São Paulo – USP, Sorocaba, São Paulo, Brazil, 18085-784; Gregório F. Azevedo, gregorio.fornetti@estudante.ufscar.br, Department of Computer Science, Federal University of São Carlos – UFSCar, Sorocaba, São Paulo, Brazil, 18052-780; Matheus V. V. Berto, matheus.berto@estudante.ufscar.br, Department of Computer Science, Federal University of São Carlos – UFSCar, Sorocaba, São Paulo, Brazil, 18052-780; Jean R. Rocha, jeanrodriguesrocha@estudante.ufscar.br, Department of Computer Science, Federal University of São Carlos – UFSCar, Sorocaba, São Paulo, Brazil, 18052-780; Eduardo C. Fidelis, 200843@facens.br, Department of Computer Engineering, Facens University, Sorocaba, São Paulo, Brazil, 18085-784; Matheus V. Nogueira, 210363@facens.br, Department of Computer Engineering, Facens University, Sorocaba, São Paulo, Brazil, 18085-784; Pedro H. Lisboa, 210331@facens.br, Department of Computer Engineering, Facens University, Sorocaba, São Paulo, Brazil, 18085-784; Tiago A. Almeida, talmeida@ufscar.br, Department of Computer Science, Federal University of São Carlos – UFSCar, Sorocaba, São Paulo, Brazil, 18052-780.

## LIST OF ACRONYMS

| Acronym | Definition |
| --- | --- |
| AAE | average attribute error |
| ACC | adaptive cruise control |
| ACF | aggregate channel features |
| ADAS | advanced driver assistance systems |
| ADE | average displacement error |
| AI | artificial intelligence |
| AOE | average orientation error |
| AP | access point |
| AP | average precision |
| ASE | average scale error |
| ATE | average translation error |
| AVE | average velocity error |
| AV(s) | autonomous vehicle(s) |
| BA-PTP | behavior-aware pedestrian trajectory prediction |
| BEV | bird's-eye view |
| BLE | bluetooth low energy |
| BS | background subtraction |
| C-V2X | cellular V2X |
| CFAR | constant false alarm rate |
| CFMSE | center final mean squared error |
| CMSE | center mean square error |
| CNN(s) | convolutional neural network(s) |
| D-VRU(s) | disabled vulnerable road user(s) |
| DPM | deformable part models |
| DSRC | dedicated short-range communication |
| EA-LSS | edge-aware lift-splat-shot |
| ELPP | EARLINET LiDAR preprocessor |
| FAF | false alarms per frame |

Continued on next page

(Continued)

| | |
|---|---|
| FDE | final displacement error |
| FMCW | modulated continuous wave radar |
| FN | false negatives |
| FoV | field of view |
| FP | false positives |
| FPGA | field programmable gate array |
| Frag | track fragmentation |
| GAN(s) | generative adversarial network(s) |
| GHz | gigahertz |
| GNN(s) | graph neural network(s) |
| GNSS | global navigation satellite system |
| GPS | global positioning system |
| HIBPN | interpreted binary Petri nets |
| HMM | hidden Markov models |
| HOG | histogram of oriented gradients |
| HOTA | higher order tracking accuracy |
| ICF | integral channel features |
| IDF1 | ID F1-score |
| IDS | identity switches |
| IMU(s) | inertial measurement unit(s) |
| INS | inertial navigation systems |
| IoU | intersection over union |
| IRL | inverse reinforcement learning |
| IRS | intelligent reflecting surfaces |
| LBP | local binary pattern |
| LDCRF | latent-dynamic conditional random fields |
| LID | local intensity distribution |
| LiDAR | light detection and ranging |
| LLM(s) | large language model(s) |

Continued on next page

(Continued)

| | |
|---|---|
| LoG | Laplacian-of-Gaussian |
| LSTM | long short-term memory |
| mAP | mean average precision |
| ML | mostly lost targets |
| MLLM(s) | multimodal large language model(s) |
| MLP | multilayer perceptron |
| MOTA | multiple object tracking accuracy |
| MOTP | multiple object tracking precision |
| MP | megapixels |
| MR | miss rate |
| MT | mostly tracked targets |
| NDS | nuScenes detection score |
| OBU(s) | on-board unit(s) |
| OCS-LBP | oriented center symmetric local binary patterns |
| OS-CFAR | statistical order CFAR |
| P2V | pedestrian-to-vehicle |
| PCA | principal component analysis |
| PDS | planetary data system |
| QSN | quantile surface neural networks |
| R-CNN | region-CNN |
| RFID | radio frequency identification |
| R-FCN | regional-fast convolutional network |
| ROI | regions of interest |
| ROLISP | risk object localization and intention and suggestion prediction |
| ROS | robot operating system |
| RSU(s) | road side unit(s) |
| SAS | self-adaptive systems |
| SDDP | simulation-driven development process |
| SORT | simple online and real-time tracking |

Continued on next page

(Continued)

| | |
|---|---|
| SSD | single shot detector |
| STFT | short-time Fourier transform |
| SUMO | simulation of urban mobility |
| TN | true negatives |
| TP | true positives |
| UWB | ultra-wideband |
| V2D | vehicle-to-device |
| V2I | vehicle-to-infrastructure |
| V2N | vehicle-to-network |
| V2P | vehicle-to-pedestrian |
| V2V | vehicle-to-vehicle |
| V2X | vehicle-to-everything |
| VLM(s) | vision-language model(s) |
| VMD | variational mode decomposition |
| VRU(s) | vulnerable road user(s) |
| VTP(s) | vulnerable traffic participant(s) |
| WLAN | wireless local area network |
| WOA | whale optimization algorithm |
| YOLO | you only look once |

## Contents

## 1 INTRODUCTION

Traffic accidents worldwide have highlighted the vulnerability of a specific group of road users known as vulnerable road users (VRUs) — or, less commonly, as vulnerable traffic participants (VTPs) — that includes pedestrians, cyclists, and motorcyclists. VRUs face heightened risks in traffic environments, making studies on their behavior and safety crucial [93, 286]. Data spanning a decade in Brazil, from 2009 to 2019, indicates that VRUs constitute a significant portion of total traffic fatalities. Although pedestrian deaths decreased from 28% to 19.3%, they still account for a substantial percentage. In contrast, the percentage of cyclist deaths remained stable at 3.5%, while motorcyclist fatalities rose significantly from 16.8% to 30.2% [47]. In India, pedestrian deaths represent a worrying percentage, estimated at approximately 19% officially, but independent studies suggest it could be as high as 35% [298]. Pedestrians in China are also significantly affected, being the most frequent victims of traffic incidents. A study in Jiangsu province revealed that pedestrians represent 50% of deaths in traffic accident [340].

Globally, an estimated 1.19 million road deaths occurred in 2021, making road traffic injuries the 12th leading cause of death across all age groups. Pedestrians account for 23% of road traffic fatalities, while cyclists and users of personal micro-mobility devices, such as e-scooters, represent 6% and 3% of deaths, respectively. Furthermore, two- or three-wheeled vehicle users account for 21% of the fatalities [222]. The global cost of road traffic injuries between 2015 and 2030 is estimated to reach US$1.8 trillion [53]. These statistics underscore the urgency of researching VRUs to understand accident dynamics and causes, and to leverage the latest technologies to mitigate this problem.

Several studies have reviewed VRUs or interactions between vehicles and VRUs. Notable among them are the works by Reyes-Muñoz and Guerrero-Ibáñez [246] and Yusuf et al. [341]. The former discusses sensing technologies and algorithms for autonomous vehicles (AVs) and their interaction with VRUs but does not cover available datasets or simulation environments for VRU-related studies. The latter reviews vehicle-to-everything (V2X) technologies aimed at improving VRU safety, briefly mentioning datasets but lacking a comprehensive survey of simulation environments.

Moreover, surveys not focused exclusively on VRUs offer valuable insights relevant to this domain. For instance, Song et al. [289] review synthetic datasets for enhancing VRU detection systems. In contrast, Feng et al. [84] summarize methodologies for deep multi-modal object detection and data fusion, presenting main datasets released between 2013 and 2019. Similarly, Micko et al. [206] investigate sensors for monitoring tasks in road transportation infrastructure, and Vargas et al. [303] review sensors for AVs, considering their vulnerability to weather conditions.

This paper provides a comprehensive review of recent studies related to VRUs, addressing critical gaps identified in previous works. Our survey was conducted using an ad hoc approach. We started by searching for initial papers across major bibliographic databases (e.g., IEEE Xplore[1], Science Direct[2]), search engines (e.g., Google Scholar[3], Scopus[4]), as well as platforms that combine both types of resources (e.g., ACM Digital Library[5]). Subsequently, an empirical and interactive literature review was conducted to gather more relevant papers. All papers deemed pertinent were properly included in this work.

We analyze the communication ecosystem between vehicles and pedestrians, which can enhance the overall perception of traffic environments and prevent accidents. This communication typically involves messages about events captured by sensors such as cameras and radars. We also examine the most relevant sensors used in VRU studies, as analyzing data collected through these sensors is essential for developing new technologies or strategies to enhance road safety. Given the costliness of data collection, researchers often rely on datasets made available by others.

We collect, analyze, and present the main datasets applicable to VRU safety research. Additionally, we explore essential methods for processing sensor data, both for developing artificial intelligence (AI) solutions and for other types of studies. Furthermore, we review the key simulation tools used to simulate traffic scenarios and generate synthetic data, which are crucial for research applying machine learning techniques to VRU safety or analyzing user behavior on roads. Simulation environments are indispensable, given the risks of conducting real-world experiments involving VRUs.

Datasets, whether collected from the literature, generated through simulations, or captured in real-time traffic environments, are fundamental for detection, tracking, classification, and intention prediction tasks. These tasks play a vital role in enhancing the perception of traffic participants, anticipating behaviors, and predicting future actions. In this study, we analyze how research in the literature addresses these tasks, examining the main factors and methods applied, aiming to better cover the available approaches and solutions in the field.

To facilitate understanding and navigation of these concepts, we have developed a taxonomy related to computational systems designed for VRU safety, as summarized in Figure 1. The taxonomy is represented by a mind map that organizes concepts found in the scientific literature into four main categories (as indicated in the legend of the figure): (i) VRU detection/classification related, (ii) VRU action/behavior/intention prediction-related, (iii) input/output-related, and (iv) algorithms and architectures. This taxonomy serves as a guide and, throughout the text, we explore these key concepts in detail, closely aligning the structure of the paper with the taxonomy. In particular, each section of the paper corresponds to one or more taxonomy categories, as presented in the following.

Section 2 surveys the communication ecosystem between vehicles and pedestrians. This section relates to categories (i) and (iv) because sensor data must be transmitted to the devices where prediction algorithms are executed. Besides that, the prediction results must reach the VRUs or other traffic agents. The communication of these messages relies on an efficient communication ecosystem, which is discussed along with the various communication technologies available.

[1]IEEE Xplore. Available at https://ieeexplore.ieee.org/Xplore/home.jsp (accessed on June 24, 2025).
[2]Science Direct. Available at https://www.sciencedirect.com/ (accessed on June 24, 2025).
[3]Google Scholar. Available at https://scholar.google.com/ (accessed on June 24, 2025).
[4]Scopus . Available at https://www.elsevier.com/products/scopus (accessed on June 24, 2025).
[5]ACM Digital Library. Available at https://dl.acm.org/ (accessed on June 24, 2025).

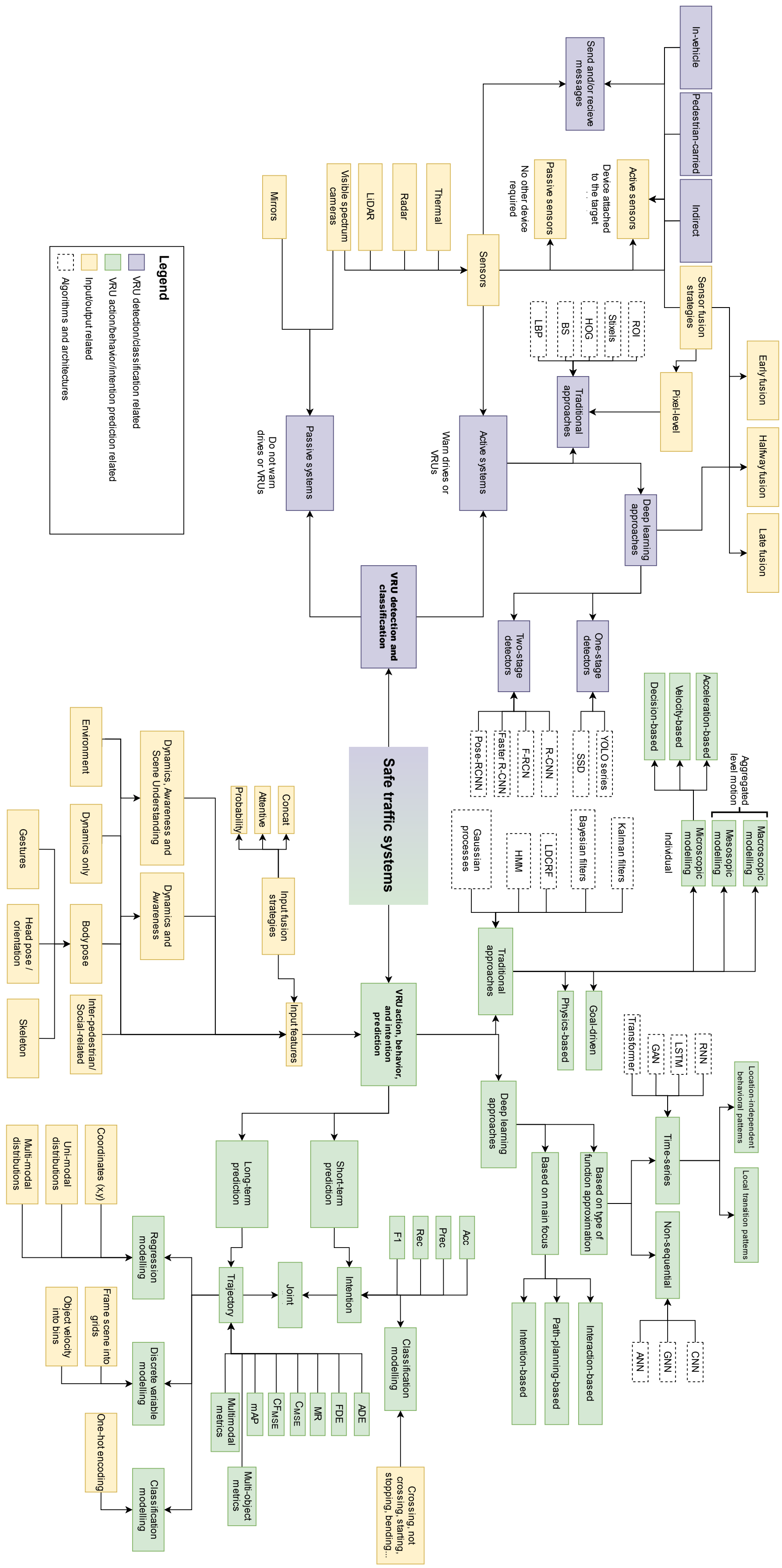

Fig. 1. VRU detection, classification, and intention prediction related taxonomy.

Section 3 presents the main types of sensors used in research on VRUs and the main datasets related to this topic. This section aligns with category (iii), as the sensors generate the input data used by the methods to enhance VRU safety. Additionally, datasets are essential for training and validating these methods.

Section 4 addresses the resources available for processing data obtained by sensors. This section relates to categories (iii) and (iv), as sensor data often undergo preprocessing to improve quality, facilitate method execution, or adapt to communication technology constraints, such as reducing data size to meet transmission capacity limits.

Section 5 focuses on simulation environments. This section aligns with category (iii), as simulation allows sensors to be virtually modeled, enabling the testing of dangerous real-world scenarios and studying potential future conditions. Simulations can also help generate datasets for training purposes. We discuss the main simulation tools, their advantages, and their challenges.

Section 6 relates to categories (i) and (iv) discussing VRU detection and classification research, detailing the algorithms employed in these processes. Section 7 relates to categories (ii) and (iv), presenting primary studies on intention prediction, behavior analysis, path forecasting and tracking, and the algorithms developed for these tasks. Finally, Section 8 summarizes the main conclusions and future work. This section ties together all taxonomy categories, reflecting on the broad concepts and insights discussed throughout the paper.

## 2 TRAFFIC ECOSYSTEM IN SMART CITIES

Enhancing the safety of VRUs within the context of smart cities demands the integration of advanced sensors, such as LiDAR (light detection and ranging) and cameras, alongside sophisticated communication technologies connecting sensors, vehicles, and VRUs. Among the most commonly utilized vehicular communication technologies is vehicle-to-vehicle (V2V) communication, which enables motorized vehicles to share real-time data, including positions, speeds, and directions.

Another pivotal technology is vehicle-to-infrastructure (V2I), facilitating the exchange of information between vehicles equipped with on-board units (OBUs) and elements of road infrastructure, known as roadside units (RSUs), such as traffic lights, cameras, and signage panels [15]. RSUs act as access points for data dissemination, mitigating the limitations of direct V2V communication [211]. Moreover, RSUs equipped with sensors can help overcome one of the major challenges in perception systems: occlusion. As highlighted by Borba et al. [41], onboard sensors often fail to detect VRUs when their view is blocked by obstacles. Infrastructure-based sensing extends the perception range beyond the vehicle's line of sight, improving the detection of occluded VRUs and enhancing safety in complex urban scenarios.

In VRU safety, vehicle-to-pedestrian (V2P) communication is central, encompassing interactions between vehicles and various types of VRUs [275]. Vehicle-to-network (V2N) communication leverages mobile networks and the internet to connect vehicles with diverse data services, providing real-time traffic conditions, weather updates, and other pertinent information from cloud services that can influence driving decisions [341]. Additionally, vehicle-to-device (V2D) technology enables direct communication between vehicles and personal devices, such as smartphones and tablets, which can be used to send alerts directly to VRUs personal devices, including proximity warnings [368]. Figure 2 provides a summary of these communication technologies.

The aforementioned types of communication (i.e., V2V, V2I, V2P, V2N, and V2D), collectively referred to as V2X, represent all forms of interaction between vehicles and various entities in the traffic environment. Cellular V2X (C-V2X), on the other hand, refers explicitly to communications technologies based on cellular network standards, such as LTE and 5G, aimed at optimizing and facilitating V2X communication [174, 341, 347].

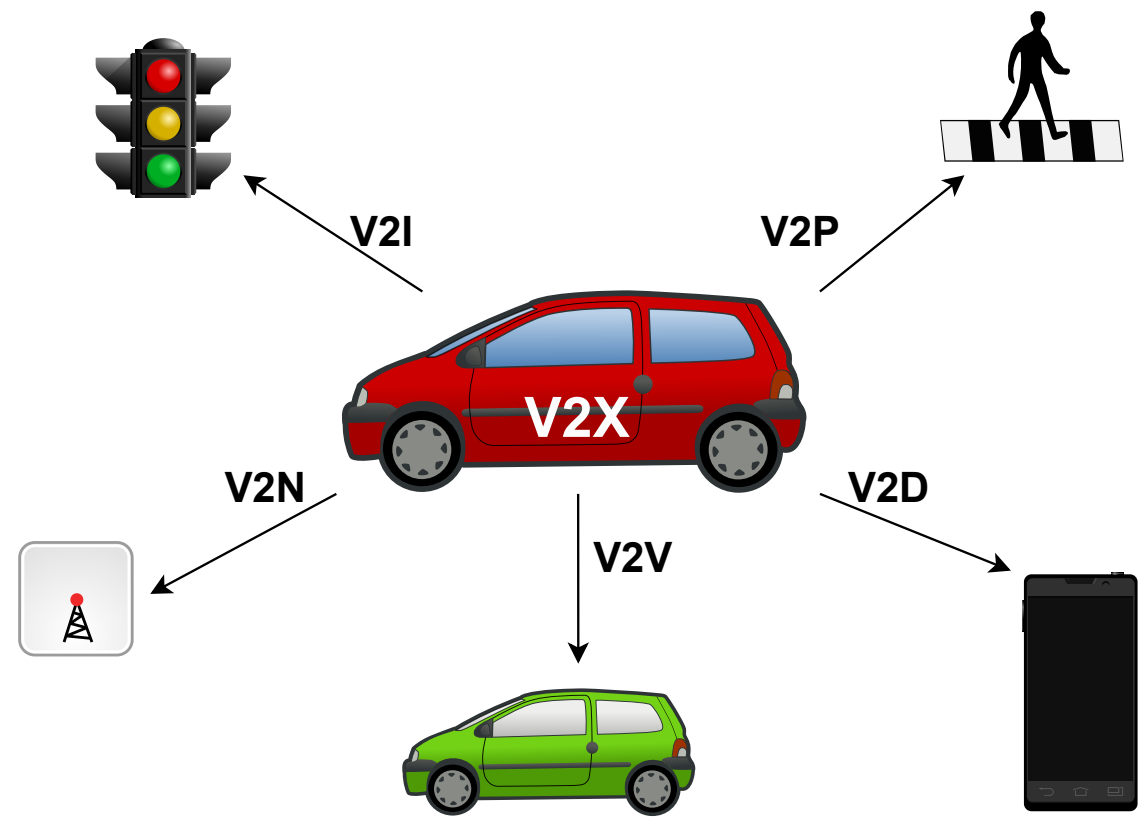

Fig. 2. Vehicle communication system.

Three primary methods are employed to implement these communications: cellular communication, Wi-Fi Direct, and dedicated short-range communication (DSRC) [293]. Wi-Fi Direct, based on the conventional Wi-Fi protocol, does not require an access point (AP) to establish connections, as one of the vehicles serves as the AP. However, this setup can introduce delays due to the additional load on the vehicle acting as the AP [142]. Conversely, DSRC communication, developed explicitly for vehicular use, offers lower latency and is considered a primary communication technology [142, 293]. Cellular technologies, such as 3G, 4G, and 5G, are also extensively used due to their advantage of not requiring specific hardware [25, 275, 293]. Table 1 provides a comparison of these communication technologies based on multiple studies [16, 67, 142, 223, 275]. Since reported values can differ between sources, representative figures were chosen to give a general idea, but actual characteristics may vary depending on different factors.

Numerous studies have explored VRU safety, incorporating both sensing and communication. Obtaining VRU positions in the environment is essential and is typically achieved using sensors and GNSS (global navigation satellite system), which includes cell phone GPS (global positioning system), often in conjunction with mobile devices like smartphones for communication [341].

For instance, Hussein et al. [135] proposed a pedestrian-to-vehicle (P2V) communication system to warn users of potential accidents, testing various communication prototypes based on 3G and WLAN. Similarly, Shahriar et al. [276] introduced a cooperative V2P method using 5G communication and GPS to alert pedestrians and drivers about possible accidents at intersections. Anaya et al. [12] investigated V2P communication for pedestrian safety via Wi-Fi, determining the minimum safe distance required between vehicles and pedestrians to issue alerts using the GPS cell phone for positioning. Another approach by Guayante et al. [111] involved using DSRC communication and multiple infrared sensors to detect VRU intentions to cross the road. Additionally, Teixeira et al. [295] employed data fusion techniques to combine information from multiple sensors within the infrastructure and GPS data to pinpoint VRU positions and issue collision warnings through communication technologies such as Wi-Fi and 5G. Figure 3 illustrates the technologies utilized in these systems.

### 2.1 Final considerations

Enhancing VRU safety within smart cities hinges on integrating advanced sensing and communication technologies. Key to this safety ecosystem is V2X communications, including V2V, V2I, V2P, and V2D, which collectively facilitate interactions between vehicles, infrastructure, and VRUs. These systems utilize various communication technologies, such as Wi-Fi Direct, DSRC, and cellular networks like 4G and 5G, each offering distinct advantages and limitations regarding latency, hardware requirements, and connectivity.

Table 1. Comparison of the communications technologies.

| Feature | Wi-Fi Direct | Cellular Communication | DSRC |
|---|---|---|---|
| **Purpose** | General-purpose communication | General-purpose communication | Specifically designed for vehicular communication |
| **Connection Type** | Peer-to-peer, no AP required | Uses existing cellular networks (3G/4G/5G) | Peer-to-peer, vehicle-to-vehicle and vehicle-to-infrastructure |
| **Hardware Dependence** | Standard Wi-Fi hardware | Employs cellular modules | Requires DSRC transceivers |
| **Coverage Area** | Short-range (up to 200m) | Wide-area (several km) | Medium-range (up to 1km) |
| **Transmission Speed** | Up to 250 Mbps | Up to 1 Gbps (5G) | Up to 27 Mbps |
| **Delay** | 1-50 ms | 50-100 ms (4G), 1-10 ms (5G) | <10 ms |
| **Power Consumption** | Moderate (depends on usage) | High (especially in mobile devices) | Low (designed for low-power vehicle applications) |
| **Advantages** | No AP required | Wide coverage, existing infrastructure | Low latency, reliability for vehicular use |
| **Disadvantages** | Potential delays if AP vehicle is overloaded | Higher latency for safety-critical apps | Requires specific hardware investment |

Studies highlight the importance of accurate VRU positioning, relying on sensors and GPS, often in conjunction with mobile devices. Several practical implementations have demonstrated the effectiveness of these technologies in preventing accidents, with systems employing communication protocols to alert drivers and pedestrians in real-time.

However, several challenges persist. One of the main issues is the high cost of implementing and effectively integrating communication technologies for collective perception of the environment, which helps reduce collisions and improve traffic flow. Ensuring low-latency communication is critical, especially in dense urban environments where interference and network congestion can impact real-time responsiveness. Although 5G and edge computing are promising solutions for this, many countries still lack widespread access to these technologies.

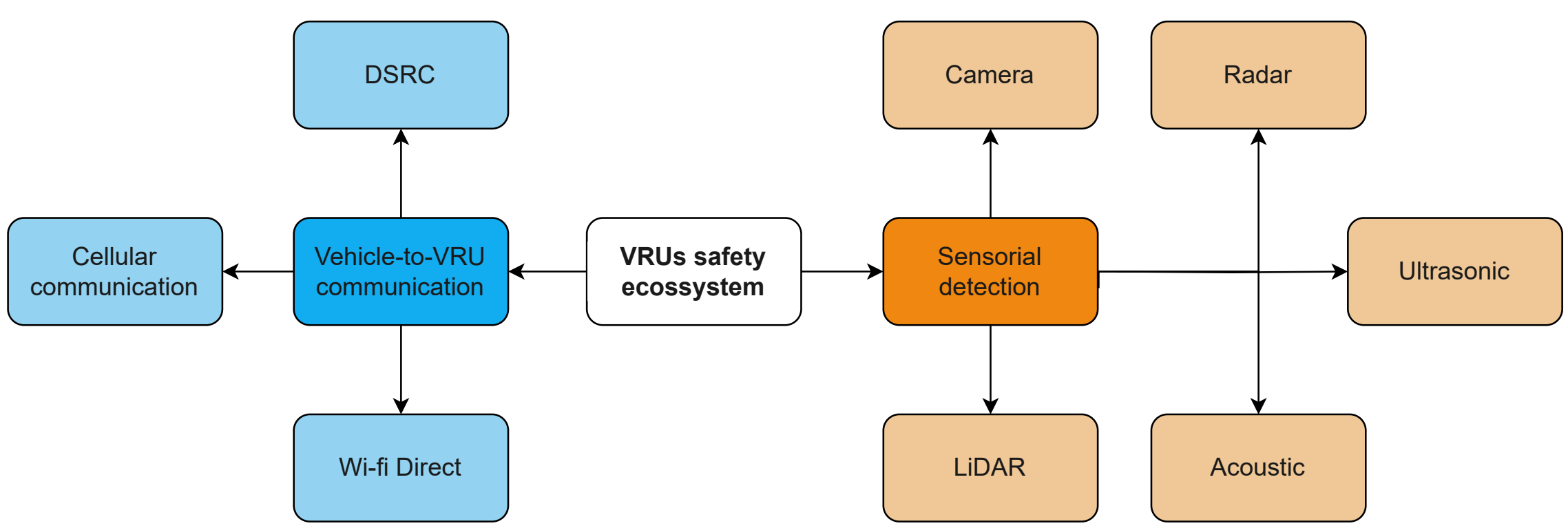

Fig. 3. Technologies involved in the VRU safety ecosystem.

Security is another critical concern. Communication technologies continue to face various privacy and security threats and vulnerabilities plaguing V2X communication. Among the solutions being explored to enhance security is the use of blockchain [250].

## 3 SENSORING AND DATA

Vehicle perception capabilities in the context of VRU detection and collision prediction rely on a variety of sensors, including cameras, LiDAR, radar, and ultra-wideband (UWB) technologies. These sensors have unique strengths and are often used in complementary ways to enhance detection accuracy and robustness in diverse environmental conditions. Table 2 summarizes the main characteristics of these sensors.

Table 2. Summary of the main characteristics of the sensors [137, 246, 303, 341].

| Feature | Visible camera | Thermal camera | Radar | LiDAR | Ultrasonic | Acoustic | UWB |
|---|---|---|---|---|---|---|---|
| Night vision capability* | Low | High | High | High | High | High | High |
| Resolution | High | Medium | Low | High | Low | Low | Medium |
| Color perception | High | Low | N/A | N/A | N/A | N/A | N/A |
| Detection range | Medium | High | High | High | Low | Medium | High |
| Field of view | Wide | Medium | Narrow | Medium | Narrow | Wide | Wide |
| Weather resistance | Low | High | High | Medium | Medium | Low | High |
| Cost | Medium | High | High | Very High | Low | Low | Medium |

* Refers to whether the sensor's captured signal or data is affected by the absence of ambient light, not its ability to generate a visual representation.

Cameras and LiDAR are known for their high resolution, which refers to the sensor's ability to differentiate nearby objects, providing detailed and dense scans of the environment. LiDAR, in particular, delivers accurate 3D information, making it a valuable tool for detailed environmental mapping [264].

Resolution can be expressed in different ways. Some sensors typically have their resolution expressed in terms of angular resolution, which is the smallest angle between two distinguishable objects. In contrast, cameras usually are expressed in spatial resolution, which refers to pixel density. The angular resolution of LiDAR can vary depending on the brand or other sensor characteristics. For instance, the angular resolution of the 16-line Velodyne LiDAR ranges from $0.1^{o}$ to $0.4^{o}$ in the horizontal direction [370]. In contrast, image resolution is typically quantified in megapixels (MP). For instance, a comparative table of cameras featured in the study by Ignatious et al. [137] reveals that resolutions range from 0.03 MP to 2.7 MP. Table 3 and Table 4 present key features of LiDAR and cameras, respectively.

Radars are generally better at detecting objects at greater distances and are highly effective in adverse weather conditions, such as heavy rain, snow, and fog, due to their longer wavelengths [261, 280]. Different types of radar, such as long-wave and microwave radar, offer distinct advantages. Long-wave radar can detect VRUs through obstacles, which is beneficial in urban environments with numerous obstructions. In contrast, microwave radar provides high-resolution data for analyzing VRU motion, speed, and distance [167]. Table 5 details key features of radars.

UWB technologies have emerged as valuable complementary sensors, particularly in scenarios where line of sight is obstructed or where traditional sensors might be compromised [126]. Table 6 highlights the main features of UWB sensors.

Ultrasonic sensors, though less common, play a significant role in enhancing VRU detection in low-speed traffic scenarios [140, 165]. Moreover, acoustic sensors have also been explored for VRU detection [100]. The main features of ultrasonic and acoustic sensors are shown in Table 7 and Table 8, respectively.

Table 3. LiDAR main features.

| Feature | Description |
|---|---|
| **Companies** | Velodyne, Hesai, Ouster, RoboSense, LeiShen, Hokuyo, IBEO, SICK [137]. |
| **Category** | Spinning mechanical LiDARs provide 360-degree coverage through rotating physical components, ideal for detailed sensing and mapping of complex environments, while solid-state LiDARs, by dispensing with moving parts, provide greater durability and reliability, although often with a more limited vision field. |
| **Beams** | LiDARs can range from a single beam to multiple beams (16, 32, 64, 128 beams are common). The greater the number of beams, the more detailed the mapping of the environment, improving detection and differentiation between objects and VRUs. |
| **Field of view (FoV)** | The FoV is the angle covered by the LiDAR sensor. It is typically described by both horizontal FoV and vertical FoV, which specify the angular range the sensor can scan. For instance, the 16-line Velodyne LiDAR, the vertical FOV ranges from -15.0$^o$ to +15.0$^o$, while the horizontal FOV is 360$^o$ [370]. |

Table 4. Camera main features.

| Feature | Description |
|---|---|
| **Category** | Visible cameras capture what we see with our eyes (reflected light), while thermal cameras detect heat signatures (infrared radiation). |
| **Frame rate** | The frame rate indicates how many images per second the camera can capture. A higher frame rate is crucial to keep up with fast movements of the VRUs. |
| **Resolution** | Camera resolution directly affects the clarity of the captured image, essential for identifying VRUs. Higher resolutions enable the detection of fine details at greater distances. |

Table 5. Radar main features.

| Feature | Description |
|---|---|
| **Frequency** | Measures the frequency of radio waves emitted by the radar. Higher frequencies offer greater resolution but less range, while lower frequencies have less resolution but better range over obstacles, such as snow, vegetation and fog. |
| **Category** | Modulated continuous wave radar (FMCW) is commonly used in vehicles to detect the distance and speed of objects. Doppler radar is used to measure the relative speed of objects. |
| **Range** | Determines how far away the radar can detect objects. The typical range for automotive radars varies and can reach up to 250 meters. |

These sensors capture crucial data at multiple stages of vehicle-VRU interaction, encompassing object detection, classification, intention prediction, and trajectory prediction [246]. Other devices like GPS, IMUs (inertial measurement units), odometers, inertial navigation systems (INS), and communication technologies (DSRC, Wi-Fi, radio frequency identification — RFID) provide critical data on vehicle positioning and dynamics, as well as the proximity of objects [206, 303]. However, this study focuses on sensors that actively emit signals or capture environmental data from vehicles or fixed urban infrastructure points for object detection and classification.

Numerous studies have provided datasets featuring data from cameras, LiDAR, and radar sensors. While many of these datasets are primarily geared toward research on AVs, they are also highly relevant for studies focused on VRU safety. The earliest datasets in this domain were predominantly camera-based. For example, the "Daimler Pedestrian Segmentation Benchmark" dataset, introduced in 2013 by Flohr et al. [89], consists of images of pedestrians manually annotated with contours. The authors captured the images using a calibrated stereo camera mounted on a vehicle navigating an urban environment.

Table 6. UWB sensors main features.

| Feature | Description |
|---|---|
| **Operation** | UWB uses ultra-wideband radio pulses to measure distances and movement with high precision, operating in a similar way to radar but with an improved ability to resolve fine details. |
| **Resolution** | Its high spatial resolution allows UWB to distinguish between objects that are very close to each other, ideal for densely populated urban environments where VRUs are often close to other objects. |
| **Resistence to interference** | UWB is highly resistant to radio frequency and multipath interference, making it ideal for congested urban environments. |

Table 7. Ultrasonic sensors main features.

| Feature | Description |
|---|---|
| **Operation** | Ultrasonic sensors work by emitting high-frequency sound waves and measuring the echo returned after these waves collide with an object. |
| **Range** | The typical range for ultrasonic sensors in vehicles varies between 0.2 and 5 meters, ideal for detecting nearby objects during parking maneuvers and in slow traffic. |
| **Field of vision** | The field of view of ultrasonic sensors is generally limited, suitable for covering specific areas around the vehicle, such as the sides and rear. |

Table 8. Acoustic sensors main features.

| Feature | Description |
|---|---|
| **Operation** | Acoustic sensors capture sound waves through microphones and use algorithms to interpret ambient sounds, identifying the presence and potentially the type of VRU based on characteristic sound signatures. |
| **Sensitivity** | High sensitivity to detect a varied range of sound frequencies, allowing for the capture of everything from pedestrian footsteps to the noise of a bicycle or motorcycle. |
| **Field of vision** | It depends on the orientation and number of microphones used. Acoustic sensors can be configured to pick up sounds from specific directions, covering areas around the vehicle. |

The KITTI dataset was released in the same year, marking a significant milestone by incorporating both camera and LiDAR data [96]. This dataset was collected using a Volkswagen station wagon with high-resolution stereo cameras, a Velodyne 3D LiDAR, and a GPS/IMU navigation system. Over six hours of diverse traffic scenarios were recorded, spanning highways to urban scenes with static and dynamic objects. The KITTI dataset includes image sequences and 3D object labels, with all data being calibrated, synchronized, and timestamped.

Datasets utilizing cameras can vary significantly based on the type of camera employed. For instance, visible cameras that capture grayscale or RGB images are used in datasets like TUD-Brussels [315]. On the other hand, thermal cameras, which capture infrared spectrum images, are used in datasets such as AITP [109].

In addition to camera and LiDAR data, several datasets incorporate radar data. Examples include nuScenes [43], radarScenes [270], REHEARSE [231], and TWICE [220]. These datasets often use real-world data, such as nuScenes [43], Waymo Open [79], and ONCE [201], to capture actual traffic conditions.

Alternatively, some datasets employ synthetic data generated through simulation tools designed to create virtual environments for testing and developing vehicle systems. Notable simulation tools include CARLA [75], SUMO (simulation of urban mobility) [31], OpenCDA [327], and CarMaker (developed by IPG Automotive). Examples of datasets utilizing these simulations are V2X-Sim [174], OPV2V [330], DOLPHINS [202], and TWICE [220].

Recently, Huang et al. [126] have proposed the WiDEVIEW dataset. This dataset stands out by incorporating traditional camera, radar, and LiDAR data, along with information collected from UWB technologies, enhancing the scope and accuracy of VRU detection and collision prediction research.

Table 9. Datasets that can be used in research on VRUs.

| **Dataset** | **Year** | **Sensor** | **Real/Simulated** | **Source** |
|---|---|---|---|---|
| UCY [163] | 2007 | camera | real | infrastructure |
| ETH [228] | 2007 | camera | real | infrastructure |
| PETS2009 [88] | 2009 | camera | real | infrastructure (surveillance) |
| TUD-Brussels [315] | 2009 | camera | real | vehicle |
| Caltech [72] | 2009 | camera | real | vehicle |
| KITTI [96] | 2013 | camera and LiDAR | real | vehicle |
| *Daimler Pedestrian* [89] | 2013 | camera | real | vehicle |
| KAIST [136] | 2015 | camera | real | vehicle |
| Tsinghua-Daimler Cyclist [171] | 2016 | camera | real | vehicle |
| CVC-14 [108] | 2016 | camera | real | vehicle |
| SDD [252] | 2016 | camera | real | drone |
| JAAD [238] | 2017 | camera | real | vehicle |
| Oxford RobotCar [194] e [27] | 2017 | camera, LiDAR, and radar | real | vehicle |
| ECP [42] | 2018 | camera | real | vehicle |
| Astyx [205] | 2019 | camera, LiDAR, and radar | real | vehicle |
| Dense[6] [110] | 2019 | camera and LiDAR | real | vehicle |
| SemanticKITTI[7] [30] | 2019 | LiDAR | real | vehicle |
| Argoverse [51] | 2019 | camera and LiDAR | real | vehicle |
| PIE [237] | 2019 | camera | real | vehicle |
| nuScenes [43] | 2020 | camera, LiDAR, and radar | real | vehicle |
| inD [40] | 2020 | camera | real | drone |
| rounD [156] | 2020 | camera | real | drone |
| BDD100K [338] | 2020 | camera | real | vehicle |
| MulRan [146] | 2020 | LiDAR and radar | real | vehicle |
| SemanticPOSS [225] | 2020 | LiDAR | real | vehicle |
| LLVIP [139] | 2021 | camera | real | infrastructure (surveillance) |
| WADS [158] | 2021 | camera and LiDAR | real | vehicle |
| BAAI-VANJEE [337] | 2021 | camera and LiDAR | real | infrastructure |
| RadarScenes [270] | 2021 | camera and radar | real | vehicle |
| Waymo Open Dataset [79] | 2021 | camera and LiDAR | real | vehicle |
| Tsinghua-Daimler Urban Pose [310] | 2021 | camera | real | vehicle |
| ONCE [201] | 2021 | camera and LiDAR | real | vehicle |
| RADIATE [280] | 2021 | camera, LiDAR, and radar | real | vehicle |

[6]Dense has more than one dataset.
[7]This dataset is based on the KITTI dataset

Table 9. Datasets that can be used in research on VRUs (continued from previous page).

| Dataset | Year | Sensor | Real/Simulated | Source |
|---|---|---|---|---|
| CODD [17] | 2021 | camera and LiDAR | simulated (CARLA) | vehicle |
| VisDrone [365] | 2021 | camera | real | drone |
| AITP [109] | 2022 | camera | real | vehicle |
| BGVP [278] | 2022 | camera | real | Internet |
| V2X-Sim [174] | 2022 | camera and LiDAR | simulated (CARLA-SUMO) | vehicle and infrastructure |
| DAIR-V2X [339] | 2022 | camera and LiDAR | real | infrastructure |
| DOLPHINS [202] | 2022 | camera and LiDAR | simulated (CARLA) | vehicle and infrastructure |
| OPV2V [330] | 2022 | camera and LiDAR | simulated (OpenCDA and CARLA) | vehicle |
| View-of-Delft [224] | 2022 | camera, LiDAR, and radar | real | vehicle |
| IPS300+ [309] | 2022 | camera and LiDAR | real | infrastructure |
| V2X-ViT [329] | 2022 | LiDAR | simulated (CARLA and OpenCDA) | vehicle and infrastructure |
| SynLiDAR [324] | 2022 | LiDAR | simulated (Unreal Engine) | vehicle |
| Deliver [351] | 2023 | camera and LiDAR | simulated (CARLA) | vehicle |
| Zenseact [9] | 2023 | camera and LiDAR | real | vehicle |
| REHEARSE [231] | 2023 | camera, LiDAR, and radar | real and simulated (synthetic rain) | vehicle |
| TWICE [220] | 2023 | camera, LiDAR, and radar | real and simulated (CarMaker) | vehicle |
| IMPTC [119] | 2023 | camera, LiDAR, and UWB | real | infrastructure |
| WiDEVIEW [126] | 2023 | camera and LiDAR | real | vehicle |
| V2V4Real [328] | 2023 | camera and LiDAR | real | vehicle |
| IAMCV [49] | 2024 | camera and LiDAR | real | vehicle |
| DAVE [312] | 2024 | camera | real | vehicle |
| SCOPE [92] | 2024 | camera and LiDAR | simulated (CARLA) | vehicle and infrastructure |
| VRUNet [281] | 2025 | camera | real | infrastructure |
| R-LiViT [208] | 2025 | camera and LiDAR | real | infrastructure |

Some datasets include real and simulated data, providing a comprehensive range of scenarios for VRU detection and collision prediction research. Notable examples are the REHEARSE and TWICE datasets [220, 231]. The REHEARSE dataset is unique as it does not rely on computational simulation tools. Instead, it simulates outdoor rainfall using rotating sprinklers to create varying intensities of precipitation within a controlled area, offering a distinct approach to data collection.

The origin of the data in the datasets also varies. Typically, sensors are installed on vehicles that traverse several kilometers, capturing interactions with other traffic participants, including vehicles and pedestrians. Prominent datasets featuring this approach include KITTI [96], nuScenes [43], and Waymo Open [79].

Some datasets employ a hybrid approach, integrating data captured from vehicles and infrastructure. In this method, sensors are mounted on RSUs, such as lamp posts and traffic signs. V2X-Sim and DOLPHINS are examples of datasets that utilize this hybrid method [174, 202]. This approach enhances collaborative perception, allowing vehicles to detect traffic participants beyond their direct line of sight by expanding their vision range [174].

Other datasets solely rely on infrastructure-based data collection. For instance, the IMPTC dataset [119] uses fixed sensors, while PETS2009 [88] and LLVIP [139] employ surveillance cameras for data acquisition.

Additionally, some datasets utilize sensors installed on drones to capture traffic data. The inD dataset [40] is an example of this approach, aiming to mitigate issues related to occlusion and behavioral changes caused by visible monitoring systems. Drones at strategic heights ensure natural user behavior and provide an aerial perspective that minimizes obstructions [40].

Certain datasets were not derived directly from sensors; instead, they were compiled using data from publicly available sources on the Internet. The BGVP dataset [278] is a notable example, consisting of manually annotated images with bounding boxes, categorized into various classes such as children, older adults, and non-vulnerable users.

Datasets relevant to VRU research often include GPS data, providing precise information on the vehicle's geographic location and the time the data is captured. This information is typically complemented by data from IMU sensors, which offer details on angular velocity and orientation [96]. Examples of datasets containing GPS and IMU information are TWICE [220], nuScenes [43], KITTI [96], WiDEVIEW [126], V2X-Sim [174], DAIR-V2X [339], OPV2V [330], V2V4Real [328], Daimler Pedestrian Segmentation Benchmark [89], and Tsinghua-Daimler Cyclist Benchmark [171]. Conversely, datasets such as REHEARSE [231], IMPTC [119], BGVP [278], inD [40], DOLPHINS [202], UrbanPose [310], ECP [42], RadarScenes [270], Waymo Open [79], and ONCE [201] do not include GPS or IMU data.

The diverse range of sensors and the comprehensive datasets available form a robust foundation for advancing VRU detection and collision prediction research. Table 9 compiles these datasets, highlighting their key characteristics, while Figure 4 maps the applications of different sensors in VRU-related studies.

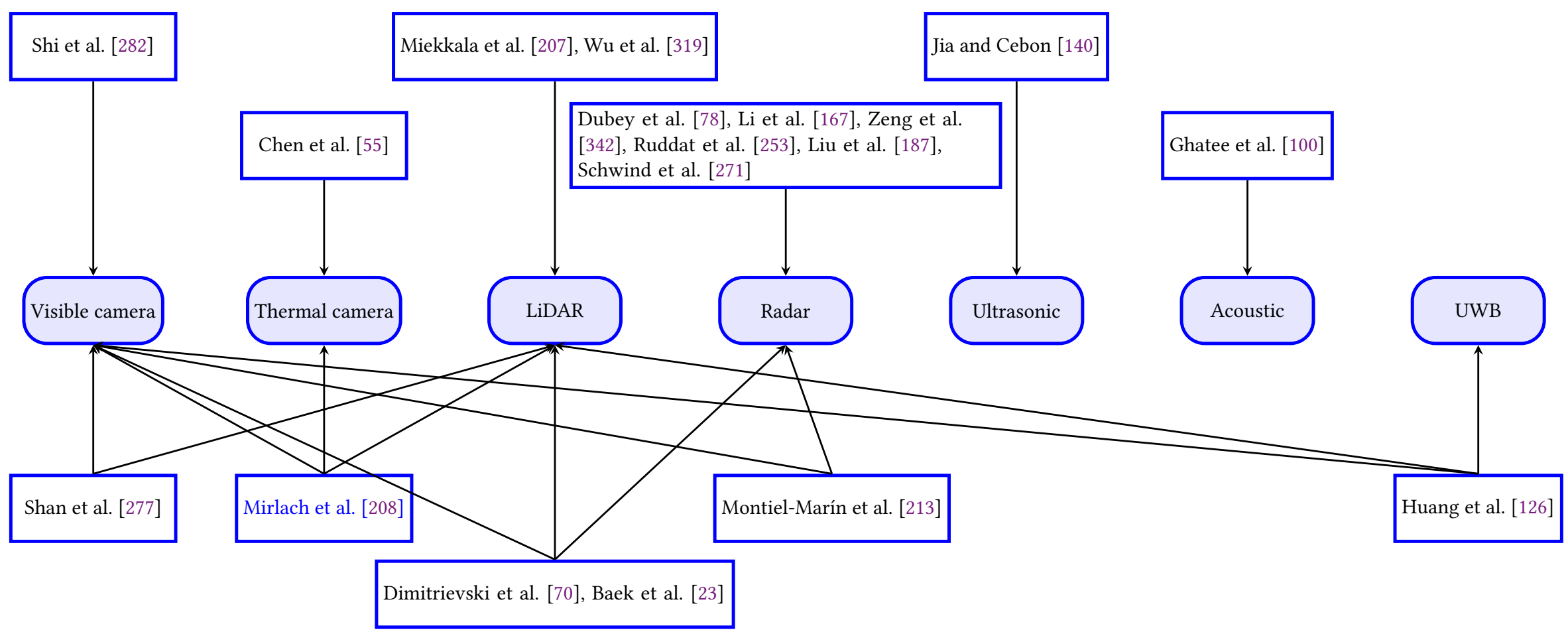

Fig. 4. Sensors applied in VRU research studies.

### 3.1 Final considerations

Although this section covered general aspects related to sensors and datasets, all these elements are directly connected to the goal of enhancing both the detection and safety of VRUs. The technologies and resources discussed are essential components for building effective solutions that can address the specific challenges faced by VRUs in modern traffic ecosystems.

The diverse characteristics of the sensors presented in this section provide the opportunity to select the one that best fits the constraints and requirements of the desired solution, allowing for adjustments in factors such as cost and infrastructure. However, the most advanced solutions today, such as those used in autonomous vehicles or in safety and monitoring systems based on RSUs, have increasingly relied on sensor fusion. This approach combines the advantages and mitigates the disadvantages of each sensor, overcoming their individual limitations to offer a more robust solution capable of handling more complex scenarios. For example, the fusion of LiDAR, radar, and thermal camera data provides a balance between detailed object detection and environmental resilience, ensuring more accurate identification of VRUs under various conditions. The following sections will explore sensor fusion in greater detail.

Additionally, we presented a comprehensive list of currently available datasets that could be useful for this sensor fusion task. The algorithms responsible for sensor fusion, VRU detection, or other tasks, require data for training and evaluation, making these datasets essential. Some of the datasets mentioned in this work contain information captured individually from vehicles or from RSUs. However, combining sensors mounted on vehicles with fixed or mobile RSUs can provide a more comprehensive understanding of VRU interactions by capturing multiple perspectives. Besides sensor fusion, it is also possible to integrate both real and simulated data.

As shown in this section, there are several datasets obtained through simulation, which expand the training possibilities for fusion methods. Simulated data offer advantages such as data augmentation and the ability to simulate dangerous scenarios that would be difficult to capture in real life, while also reducing the costs associated with generating labeled data. However, real-world data are indispensable for validating and adapting models to unpredictable environments, as simulated data alone cannot replicate the full complexity of the real world. The use of simulated environments will be further discussed in the following sections.

## 4 PREPROCESSING DATA TECHNIQUES

Preprocessing of sensor data is fundamental to guaranteeing the quality and usefulness of the collected information. This process involves removing noise, correcting errors, and extracting relevant characteristics from the raw data, resulting in more accurate and effective analysis [294].

The initial step is often noise reduction, which is important for enhancing data quality. Techniques such as Gaussian filtering, median filtering, and wavelet transforms are commonly used to reduce noise in image data [2]. For LiDAR data, methods like statistical outlier removal and radius outlier removal are employed [322]. Due to its susceptibility to various noise sources, radar data often requires advanced filtering techniques like Kalman filtering and clutter removal [91].

Feature extraction is a critical step that involves identifying and isolating relevant characteristics from the data. This can include edge detection, texture analysis, and color space transformations for image data. LiDAR data preprocessing often involves extracting geometric features such as points, lines, and planes, which are essential for object detection and classification. Radar data features typically include velocity, range, and angle of arrival, which provide valuable information for tracking and identifying objects [341].

In VRU detection, image normalization is a common preprocessing technique that standardizes data fed into deep learning models, enhancing their generalization capabilities [200]. Normalization adjusts the pixel intensity values to a common scale, which is necessary for the consistent performance of neural networks.

Data augmentation is a technique used to expand the dataset by applying transformations, such as rotation, scaling, mirroring, and cropping, to the original data. This practice enhances the robustness and generalization of machine learning models by exposing them to various scenarios. Augmentation is particularly beneficial in addressing the issue of limited training data, which is common in VRU detection tasks [62, 341].

Sensor fusion combines data from multiple sensors to leverage their complementary strengths, resulting in more robust VRU detection [287]. Techniques proposed by Aziz et al. [21] demonstrate the benefits of sensor fusion, such as combining radar and camera data to compensate for the individual weaknesses of each sensor. Preprocessing for sensor fusion involves spatial and temporal data synchronization from different sensors. Spatial calibration ensures that data points from different sensors align correctly in the same coordinate system, while temporal calibration ensures that data points are synchronized in time. This is critical for accurately associating data points from different sensors and subsequent tasks such as object tracking and classification.

Advanced techniques are continually evolving to meet the challenges of VRU detection in complex environments. For instance, deep learning-based denoising methods are being developed to improve noise reduction in image and LiDAR data [341]. Additionally, real-time data processing techniques are becoming increasingly important for applications in autonomous driving, where timely and accurate VRU detection is critical.

In the following, we present the main preprocessing techniques used to manipulate data collected from radars, LiDAR, and cameras.

### 4.1 Radar data

Radar sensors play a central role in VRU detection by capturing detailed environmental information, even under adverse weather and lighting conditions, making them ideal for advanced driver assistance systems (ADAS). However, raw radar data often requires preprocessing to remove noise, correct distortions, and extract relevant features for effective detection and classification of VRUs [300].

Gamba [91] provides an in-depth review of radar signal processing for autonomous vehicle applications. They emphasize the importance of Fourier transforms in converting signals from the time domain to the frequency domain, facilitating efficient analysis and the application of various signal processing algorithms. Additionally, Ullmann et al. [300] discuss applying filters to remove static noise and using the short-time Fourier transform (STFT) to obtain micro-Doppler signatures.

Liu et al. [184] discuss the preprocessing of radar data from the Tianwen-1 rover, emphasizing the conversion of raw data into the planetary data system (PDS) format. Techniques include phase and time calibration, background removal, and gain adjustment to improve radar image accuracy and clarity. Despite its extraterrestrial focus, we can apply these methodologies to terrestrial radar systems that monitor VRUs.

Li et al. [176] demonstrate how radar point cloud projection on the image plane combines sparse radar data with visual information to improve 2D and 3D object detection. The CenterTransFuser model uses a fusion approach that processes radar data and RGB images independently before combining them into a cross-transformer module, increasing detection accuracy for pedestrians, motorcycles, and bicycles. Scheiner et al. [261] address the sparsity of radar data by accumulating radar points over multiple timestamps to create a denser representation, though this approach must manage additional noise.

Other studies employed CFAR (constant false alarm rate), a target detection technique widely used in radar signal processing, especially in environments with uncertain or variable noise [195]. Kong et al. [151] propose a two-level preprocessing algorithm based on combined CFAR, which aims to improve object detection using 4D radar. The algorithm initially applies a coarse CFAR with a relatively high threshold to remove low-power noise measurements. They apply a ordered statistic CFAR (OS-CFAR) with a lower threshold to the measurements preserved along the azimuth axis to minimize invalid measurements and produce reliable and valid measurements.

Several studies highlight deep learning methods to enhance radar signal processing. González [107] presents systems that classify VRUs based on single-frame radar measurements using convolutional neural networks (CNNs) and approaches that extract regions of interest from the Range-Doppler spectrum for classification using you only look once (YOLO). Cha et al. [50] explore preprocessing FMCW radar sensor data by converting raw signals into Range-Doppler maps and point cloud maps, which we can employ as input for a deep learning architecture based on CNNs.

Table 10 summarizes the main radar data preprocessing methods.

Table 10. Main radar data preprocessing methods.

| Method | Pros | Cons |
|---|---|---|
| Fourier Transform | Facilitates the conversion of signals from time domain to frequency domain, enhancing signal analysis [91]. | Complex understanding required; may not handle non-linear or non-stationary signals effectively [300]. |
| Static Noise Filtering | Effectively removes low-power noise measurements, improving data clarity [184]. | Risk of eliminating weak but significant signals during initial high threshold filtering stages [151]. |
| Statistical CFAR | Improves detection reliability and adapts to different radar scene dynamics [195]. | Computationally intensive; requires tuning based on specific settings [151]. |
| Deep Learning | Improves traditional steps like target detection; utilizes raw signals effectively [107]. | Needs extensive training data and high computational resources [50]. |

### 4.2 LiDAR data

LiDAR sensors capture detailed three-dimensional environmental information, generating point clouds that accurately represent objects and surfaces. These sensors efficiently obtain road measurement data and assess road conditions, playing a critical role in VRU detection [325] and the development of intelligent transportation systems [141].

Raw waveform LiDAR data typically exhibits extended, misaligned, and relatively detail-free features, requiring preprocessing to ensure data quality and accuracy. Wu et al. [322] address these issues by applying a preprocessing chain that includes frequency-based noise filtering, Richardson-Lucy deconvolution, waveform registration, and angular rectification. This method was validated using high-fidelity simulations, demonstrating significant improvements in waveform signal recovery.

Noise reduction is a critical step in preprocessing LiDAR data. Li et al. [166] combine variational mode decomposition (VMD) with the whale optimization algorithm (WOA) to reduce noise in LiDAR signals. The proposed method optimizes VMD decomposition parameters, using the Bhattacharyya distance to identify relevant modes for reconstruction. The result is a higher signal-to-noise ratio and extended detection range.

For processing raw LiDAR data, D'Amico et al. [65] developed the EARLINET LiDAR pre-processor (ELPP), an open-source module that performs instrumental corrections and data manipulation of raw LiDAR signals. ELPP automates tasks such as dead time corrections, background subtraction, and signal smoothing, providing statistical uncertainties through error propagation or Monte Carlo simulations.

Zhou et al. [360] propose an improved Gaussian decomposition method for LiDAR echoes, implemented on a field-programmable gate array (FPGA) to enhance processing speed and accuracy. This method is validated using LiDAR datasets from the Congo and Antarctic regions, demonstrating significant improvements in processing efficiency.

In point cloud processing, Duan et al. [77] present an adaptive noise reduction method based on principal component analysis (PCA), reducing computational complexity while maintaining environmental feature details. Other studies, such as Passalacqua et al. [227], explore the extraction of channel networks from LiDAR data to improve object segmentation and VRU identification in urban environments. Zhao et al. [358] explore multi-task learning networks for preprocessing complex LiDAR data.

Table 11 summarizes the main LiDAR data preprocessing methods.

Table 11. Main LiDAR data preprocessing methods for VRU detection.

| Method | Pros | Cons |
|---|---|---|
| Frequency-Based Noise Filtering | Improves clarity by removing frequency-specific noise, enhancing signal accuracy [322]. | May not effectively isolate non-frequency specific distortions. |
| Richardson-Lucy Deconvolution | Enhances resolution by correcting blurring effects, facilitating better object delineation [322]. | Computationally intensive; can amplify noise if not properly tuned. |
| Variational Mode Decomposition | Enables refined decomposition of signal components, improving identification of relevant LiDAR echoes [166]. | Parameter tuning is critical and can be complex to optimize. |
| ELPP | Automatically corrects and manipulates raw signals for advanced optical processing [65]. | Specific to aerosol data; may need adjustments for other types of LiDAR applications. |
| Gaussian Decomposition for FPGA | Significantly faster processing suitable for real-time applications, maintaining high accuracy [360]. | Requires FPGA hardware; may not be as flexible as software solutions. |
| Adaptive Noise Reduction via PCA | Reduces noise while preserving detail, reducing computational load [77]. | PCA-based method may struggle with highly irregular or sparse data sets. |

### 4.3 Camera data

Preprocessing camera data is crucial for ensuring the quality and usefulness of images used in computer vision. Techniques such as distortion correction, lighting adjustment, normalization, and noise removal are essential for improving the accuracy of pattern recognition algorithms. With the advent of CNNs and other deep learning models, the focus has shifted to data augmentation, creating variations of training data to enhance model robustness and generalization [62]. This shift is due to the ability of deep learning models to automatically discover and apply filters and extract high-level features from the images. Despite this, some research indicates that traditional methods remain important, as handcrafted features can be effectively combined with those discovered by deep learning methods. This hybrid approach can enhance the overall performance of VRU detection and classification systems [296].

Murcia-Gómez et al. [217] highlight the importance of lighting correction and contrast enhancement to mitigate lighting variations between images, which is essential in traffic environments. Filters such as exponential, gradient, Laplacian-of-Gaussian (LoG), local binary pattern (LBP), logarithmic, square, square-root, and wavelet filters are commonly used for image preprocessing Demircioğlu [69].

Abuya et al. [2] provide an overview of image processing filters like Gaussian, Sobel, median, Laplacian, and average filters, which improve image quality across various domains, including VRU detection. These techniques can significantly enhance the accuracy and reliability of data in traffic-related tasks.

For feature extraction, techniques like histogram of oriented gradients (HOG) are popular in pedestrian detection approaches [46]. Dollár et al. [73] integrate integral channel features (ICF), aggregate channel features (ACF), and deformable part models (DPM) within the fast feature pyramids framework, demonstrating their effectiveness in extracting discriminative features for object detection.

Color transformations (e.g., converting to the LUV color space) are often employed in VRU detection. This space separates the luminance from color components, allowing algorithms to treat lighting and colors independently, enhancing detection effectiveness [296]. Zhu and Yin [367] use LAB color space for shadow detection and removal in autonomous vehicles, further highlighting the importance of color-based preprocessing.

Thermal cameras offer additional preprocessing challenges and opportunities [313]. Techniques such as local intensity distribution (LID), oriented center symmetric local binary patterns (OCS-LBP), and histograms of regions of interest (ROI) are commonly used for feature extraction in thermal images, enhancing VRU detection and classification [183].

Table 12. Main camera data preprocessing methods for VRU detection.

| Method | Pros | Cons |
|---|---|---|
| Distortion Correction | Improves geometric accuracy of images, essential for precise measurements. | Requires accurate camera calibration; computationally intensive. |
| Lighting Adjustment | Mitigates lighting variations, enhancing image consistency. | May introduce artifacts if not applied carefully; varies with lighting conditions. |
| Normalization | Standardizes pixel intensity values, improving algorithm performance. | Can reduce contrast if not tuned correctly. |
| Noise Removal | Enhances image clarity, important for accurate pattern recognition. | Risk of losing important details if over-applied. |
| Data Augmentation | Enhances model robustness and generalization by creating training data variations. | Requires extensive computational resources; can lead to overfitting if not balanced. |
| Handcrafted Features | Combining with deep learning features enhances performance. | May require significant domain knowledge and tuning. |
| Color Transformations | Separates luminance and color components, improving detection effectiveness. | May complicate processing pipeline; effectiveness varies by application. |
| Image Filters | Improve image quality through various methods (Gaussian, Sobel, median, etc.). | Each filter has specific limitations; may require multiple filters for best results. |
| Thermal Image preprocessing | Enhances feature extraction in thermal images, necessary for VRU detection. | Complex and requires specialized techniques. |

Table 12 summarizes the main camera data preprocessing methods.

### 4.4 Final considerations

The techniques discussed in this section, such as noise reduction, feature extraction, and data fusion, are broadly applicable to sensor data, but they also play a critical role in enhancing both detection and safety of VRUs. Throughout the section, we highlighted applications and references directly related to VRUs, reinforcing how these general preprocessing strategies are crucial for developing robust systems for VRU scenarios.

Preprocessing sensor data is essential to ensure the quality and usefulness of the information collected, allowing for more accurate and effective analysis [294]. This process usually involves reducing noise, correcting errors and extracting relevant features.

Noise reduction techniques in image data include methods such as Gaussian filtering, median filtering, and wavelet transforms [2]. For LiDAR data, methods like statistical outlier removal and radius outlier removal eliminate erroneous points in point clouds [322]. Moreover, advanced filtering techniques such as Kalman filtering and CFAR are applied to radar data to mitigate noise and clutter, enhancing signal quality [91, 151].

Feature extraction is a fundamental step that helps to identify important aspects of the data, such as texture and edges for images [296], as well as geometric features in LiDAR data and range and speed parameters in radar data [341]. In addition, image normalization is commonly applied in detection tasks, improving the generalization capacity of deep learning models and the accuracy in delimiting regions of interest [200]. Also worth mentioning the importance of data augmentation, which enhances model robustness and addresses limited training data issues common in VRU detection tasks by applying transformations such as rotation, scaling, mirroring, and cropping [62].

Sensor fusion, another critical aspect, combines data from different types of sensors to compensate for individual limitations, using spatial and temporal calibration for proper data alignment [287]. Other advanced techniques, such as deep learning-based ones, are also explored and continue to evolve to meet the needs of complex environments, especially in the detection of VRUs on urban roads and in ADAS [341].

## 5 SIMULATION ENVIRONMENTS

Simulation environments are indispensable to advancing VRU detection and collision prediction research. These environments facilitate the understanding and identification of critical situations affecting the safety of traffic users by enabling the creation and application of models tailored to specific research and experimentation objectives. Although we begin this section with a general overview of simulation tools and their capabilities, our primary focus is on how these tools are used in VRU-related tasks, which is reflected in the majority of the studies and applications discussed below.

Several tools are key for vehicle simulation and testing. Notably, CARLA [75], SUMO [31], OpenCDA [327], and CarMaker [162, 299] are widely used. CARLA and SUMO are open-source platforms, while OpenCDA is freely available for academic research. Other frequently utilized tools include Unity 3D and OMNet++ [302]. These platforms have benefited from technological advancements, allowing the replication of real-world data in simulation environments.

Microsimulation models such as SUMO and OMNeT++ enable the simulation of individual behaviors within a road network and across an entire city's traffic system. These tools offer detailed 2D representations of the road environment [190, 247]. They are important for creating scenarios that include various types of VRUs, providing accurate and comprehensive traffic simulations.

Integrating multiple simulation tools can create more complex and realistic scenarios. For example, the SUMMIT simulator, developed by Cai et al. [45] as an extension of CARLA, utilizes OpenStreetMap data to generate intricate urban environments. This integration inherits CARLA's physics and visual realism, facilitating the testing of driving algorithms in dense, unregulated urban settings.

Three-dimensional simulation environments, including game engines like Unity 3D [133], Unreal Engine 4 [75], and platforms like CARLA, offer advanced visual realism and physics necessary for autonomous driving simulations. These environments simulate interactions among vehicles, VRUs, and infrastructure at different levels of road networks. This approach enables comprehensive studies on user behavior and the development of applications that enhance traffic safety [354].

Some simulation environments, such as CARLA and Unreal Engine 4, are capable of simulating multiple sensors commonly used in VRU research, such as cameras, LiDAR, radar, IMU, and GPS, and even provide photorealistic rendering under varied conditions (e.g., lighting, weather, occlusion). However, despite these advances, discrepancies remain between synthetic and real-world sensor returns, particularly in aspects such as texture, material reflectivity, and noise characteristics. Furthermore, differences in vehicle dynamics and agent behavior generated by these tools can also affect the realism and reliability of the simulations [273]. This gap critically affects VRU-detection and safety algorithms: for example, LiDAR-based object detectors trained solely on simulated data often underperform in real-world scenes [131]. Bridging this simulation-to-real gap therefore demands techniques such as careful sensor calibration, domain adaptation techniques, training pipelines that combine real and synthetic data, and the use of digital twins in simulators such as CARLA to better align simulated and real-world behaviors [273].

Some research extends beyond traditional simulation tools, allowing direct human interaction with the simulated environment. For instance, Artal-Villa et al. [18] developed a 3D driving simulator on Unity, integrating pedestrians and other vehicles. This simulator used data from SUMO, enhancing the accuracy of interactions between traffic elements and contributing to detailed road safety analyses for VRUs.

Complementing this perspective, Gueye et al. [113], proposes a pedestrian decision model for illegal road crossing, motivated by the lack of pedestrian crosswalks in cities such as Dakar. The model is designed to reflect the traffic behavior typical of urban environments in developing countries. The authors evaluate their approach through three simulation scenarios using SUMO, leveraging its microscopic traffic flow modeling capabilities to simulate realistic interactions between pedestrians and vehicles. This further illustrates that, although traffic flow simulators such as SUMO are primarily designed for broader mobility studies, they are highly applicable to VRU-focused research.

In addition to traffic flow simulators like SUMO, several studies explore more advanced simulation platforms. For instance, Gómez-Huélamo et al. [117] validated an autonomous driving architecture using the robot operating system (ROS) within the CARLA simulator, emphasizing decision-making in complex urban scenarios. This study employed hierarchical interpreted binary Petri nets (HIBPN) to manage dynamic situations involving VRUs, focusing on scenarios like pedestrian crossings and adaptive cruise control (ACC). Similarly, Won and Kim [318] proposed a simulation-driven development process (SDDP) using CARLA, focusing on VRU safety. By implementing Euro NCAP (European New Car Assessment Programme) test scenarios through the ASAM (Association for Standardization of Automation and Measuring Systems) OpenSCENARIO format, the study validated autonomous vehicle system requirements and optimized values for ADAS.

Keler et al. [145] used the SUMO simulation environment to model interactions between AVs and VRUs at urban roundabouts. The study leveraged real observational data to simulate and analyze these interactions, defining maneuver classes and driving strategies to study explicit and implicit communications between VRUs and AVs.

In addition to the widely recognized simulators, tools like VISSIM are also extensively used. VISSIM is a detailed microsimulation environment capable of replicating real-world conditions, providing complex vehicle and pedestrian behavior analyses across different road networks. For instance, combining VISSIM with PC-Crash for ADAS and VRU safety development has shown that object visibility and reaction time significantly impact active safety systems' effectiveness [150].

Figure 5 presents a comprehensive overview of the most commonly used simulators and tools in traffic safety research focused on VRUs. It highlights the fundamental studies that have employed these simulation methods, showcasing the breadth and depth of simulation-based research in enhancing VRU safety.

### 5.1 Final considerations

The simulation tools discussed in this section play an essential role in research and development aimed at VRU safety, offering different features and capabilities depending on the study objectives. They vary in terms of realism, data generation, and computational requirements, which directly impact the choice of the most suitable simulator.

CARLA is notable for its realistic 3D simulations, ideal for autonomous vehicle algorithms in complex urban scenarios [75], while SUMO focuses on large-scale traffic modeling [31, 113, 299]. OpenCDA combines the capabilities of CARLA and SUMO, making it more suitable for testing connected and autonomous vehicles in specific scenarios [327].

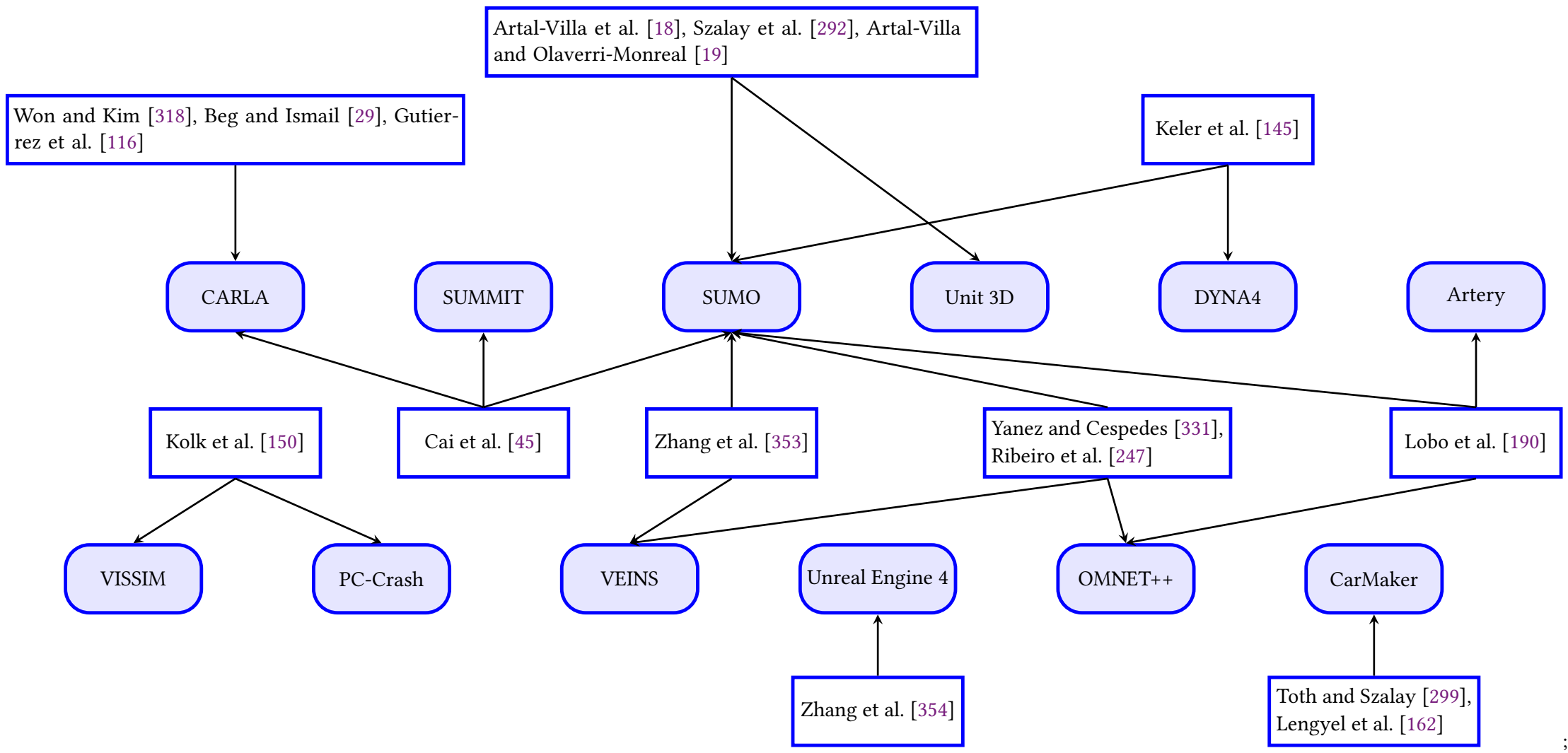

Fig. 5. Main simulators used in VRU research.

Tools like VISSIM offer high precision in microsimulations of interactions between vehicles and VRUs, although they have cost and licensing restrictions [150]. Game engines, such as Unity 3D and Unreal Engine, provide flexibility to create customized scenarios and develop ADAS interfaces, but face challenges in integration with traffic models [133].

OMNeT++ is effective in modeling communication networks at a microscopic level, enhancing urban traffic analysis when integrated with SUMO [302]. In turn, SUMMIT integrates OpenStreetMap data into CARLA, enabling realistic simulations of unregulated urban environments, offering a more holistic approach to VRU safety [45].

Each simulator has its strengths in different application contexts. For example, VISSIM is suitable for detailed interactions, SUMO is more appropriate for large-scale traffic analysis, while CARLA and SUMMIT are preferred for autonomous driving algorithms in complex environments. Table 13 summarizes these characteristics, highlighting their advantages, disadvantages, and suitability for different scenarios.

Among the challenges associated with simulation tools is the difficulty in accurately modeling the behavior of various traffic agents in urban environments, such as the simulation of unpredictable pedestrian movements. Additionally, the visual or data quality generated in simulations often falls short of real-world conditions, complicating the use of simulated data for training algorithms intended for real-world applications, such as VRU detection. Moreover, these tools tend to be computationally intensive, requiring substantial resources.

## 6 VRU DETECTION AND CLASSIFICATION

Traditional vehicles rely on components such as mirrors and conventional cameras to assist drivers in recognizing VRUs or potential road hazards. However, these elements act as passive systems and cannot alert drivers to potential blind-spot accidents. In contrast, active systems can warn drivers or VRUs of imminent collisions and have shown to positively influence, for example, drivers' yielding rates — the frequency with which they yield or give way to other road users [134].

Table 13. Comparative analysis of simulation tools for VRU safety research

| Simulator | Pros | Cons | Scenarios |
|---|---|---|---|
| CARLA | High visual and physical realism; ideal for autonomous driving algorithms. | Requires high computational power. | Complex urban scenarios. |
| SUMO | Large-scale traffic modeling; robust quantitative analysis. | Limited visual realism. | Traffic flow studies. |
| OpenCDA | Combines CARLA and SUMO features; focused on connected vehicles. | Restricted to connected and autonomous vehicle scenarios. | Connected and autonomous vehicle scenarios. |
| VISSIM | High precision in vehicle-VRU interactions. | High costs and licensing restrictions. | Detailed microsimulations of interactions. |
| Unity 3D | Versatility for customized scenarios and ADAS interface development. | Challenges in integration with traffic models. | ADAS interface development and customized scenarios. |
| OMNeT++ | Effective in modeling communication networks at a microscopic level. | Limited visual elements. | Vehicle communication simulation. |
| SUMMIT | Integration with OpenStreetMap; realistic simulations of unregulated environments. | Limited to unregulated urban scenarios. | Testing in unregulated environments. |

Moreover, in automotive applications, like fully AVs and ADAS, active systems provide warnings and take proactive measures to ensure the safety of all road users. The ability to perceive VRUs and nearby objects is vital, given that the decision-making processes of AVs and ADAS heavily rely on real-time data extracted from the traffic environment. Given that, computational systems dedicated to VRU safety focus on preventing accidents involving pedestrians and cyclists. Over the past decades, there has been a notable surge in VRU detection and classification studies. Figure 6 illustrates the increasing volume of papers on these subjects published in the IEEE Xplore Digital Library[8].

Researchers have proposed several methods for detecting VRUs in traffic environments, each aiming to enhance results and mitigate individual limitations inherent in the employed sensors. These limitations may stem from occlusion [200], data resolution [33, 262], sensitivity to illumination [90], weather conditions [229], temperature levels [272], implementation cost [5], and velocity measurement [341].

One of the most effective approaches to overcoming these limitations is sensor fusion. By integrating multiple sensor types with complementary strengths, fusion techniques help mitigate issues such as occlusion, limited line of sight, sensitivity to illumination, weather conditions, and velocity measurement errors [124, 336]. For instance, as shown in Table 2, visible cameras provide high-resolution color images but struggle in low-light conditions, while thermal cameras and radar maintain performance regardless of illumination. Similarly, LiDAR offers high-resolution spatial data but can be affected by adverse weather, whereas radar and UWB sensors remain robust under such conditions. In addition, cameras may face limitations in velocity measurement due to frame rate constraints, which can affect their performance in certain scenarios. However, they excel at capturing intricate visual details such as color, contours, and texture. In contrast, radar sensors directly measure the relative velocity of objects using the Doppler effect, but can struggle to differentiate between nearby VRUs, especially in dense environments.

[8]The results were retrieved using the query `(("All Metadata":"vulnerable road user") OR ("All Metadata":"VRU")) AND (("All Metadata":"detection") OR ("All Metadata":"classification"))`.

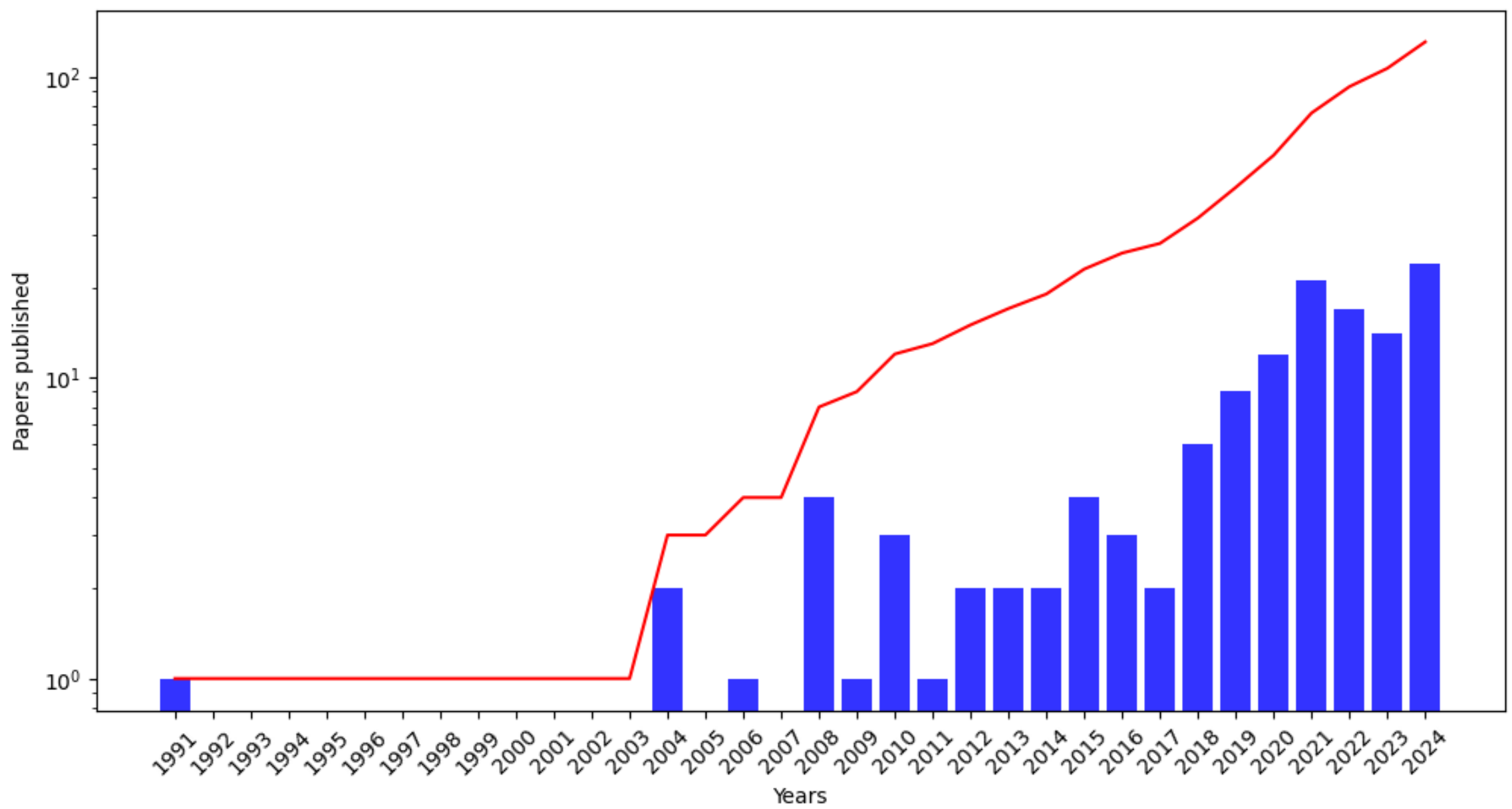

Fig. 6. Studies related to VRU detection or classification available in IEEE Xplore by year. The red line indicates the cumulative sum.

With sensor fusion, detection models can combine the benefits of different technologies to enhance overall reliability. For example, the combination of camera and radar through fusion techniques can enhance velocity estimation from cameras and improve object separation from radar, enabling reliable object detection and tracking in complex environments [24].

Datasets, such as those mentioned in Section 3 (e.g., REHEARSE and TWICE), are also essential for overcoming the limitations of certain sensors. These datasets are important for training models to account for variations in environmental conditions, thereby ensuring robustness and reliability across a wide range of real-world scenarios.

Individual sensor limitations can also be mitigated through V2X applications, particularly cooperative awareness and collective perception. A V2X infrastructure allows sensors from multiple vehicles to share data, while vehicles themselves can transmit their current position, speed, and heading. This shared information can be used by automated detection and tracking systems, enhancing tasks such as trajectory prediction and improving robustness against challenges such as occlusion and limited line of sight [263, 317].

In addition to fusion and collective perception, some limitations can also be addressed through preprocessing techniques. For instance, image processing methods can help mitigate the effects of low-light conditions and reduce noise, as discussed in Section 4. In addition, machine learning-based enhancement techniques, often deep learning approaches, offer effective solutions. Super-resolution methods, for example, can reconstruct finer details in low-resolution images or LiDAR data, improving overall detection accuracy and reliability [123, 323].

We can categorize the majority of the classification solutions into three main groups: (i) in-vehicle devices, (ii) pedestrian-carried devices, and (iii) indirect systems. The first group comprises intelligent systems integrated within the vehicle structure, automatically identifying risky situations on the road. For instance, Alaqeel et al. [7] demonstrated how the human body exhibits distinct responses to millimeter-wave radars — recently regarded as mature and affordable technologies — at J-band frequencies (220 to 325 GHz) compared to vehicles. Therefore, the study conducted a phenomenological experiment that establishes a foundation for AVs with identification algorithms leveraging the analyzed features and supports potential regulations for these sensors and frequencies. Similarly, Dubey et al. [78] used simulated radar data to train convolutional and long short-term memory (LSTM) networks to classify VRUs as pedestrians or bicyclists, reaching an accuracy of 99.81% with a combination of those models. They proposed a four-stage radar signal processing pipeline with short-time Fourier transformations (STFT) to classify targets with a wide range of velocity components, typically observed in VRUs. However, in-vehicle strategies often exhibit limitations such as short-range operation capability or low noise robustness at high speeds [111].

The pedestrian-carried devices approach assumes individuals carry an object (e.g., smartphone or smartwatch) that transmits input data to vehicle sensors. Although not all gadgets are designed for AVs, they demonstrate potential for high mobility support, high bit-rate communication range, and capacity [14]. Smartphone-based technologies can also be seamlessly integrated with cloud services [25], achieving reasonable results in V2P communications [14, 179, 291]. For example, Verhaevert [306] proposed a system using off-the-shelf products with bluetooth low energy (BLE) capabilities — commonly present in smartphones [355] — to detect VRUs in truck blind spots. The authors demonstrated that the BLE devices could function effectively with a package transmission interval of 20ms, making them suitable for real-time detection tasks.

Some studies have developed specific pedestrian-carried device prototypes for communication with connected vehicles. This strategy requires attaching hardware to the objects to be detected, known as active sensors [5]. For example, Zhang et al. [347] also leveraged the common occurrence of pedestrians carrying smartphones and introduced a VRU warning system based on a phone case for V2P communication. The method uses a GNSS to share information with nearby connected cars, which alert drivers via sounds, icons, or vibrations. Experiments conducted with this system in various scenarios demonstrated a distance range of 200 m and a positioning accuracy of 0.83 m. Additionally, Lazaro et al. [160] developed and evaluated a proof-of-concept tag for scooters or bicycles to broadcast millimeter waves detectable by radar sensors on AVs. Their method employs semi-passive millimeter wave identification (MMID) to reduce clutter within urban environments without specific lanes for two-wheeled users. Furthermore, this tag and other radar-based solutions could be enhanced using intelligent reflecting surfaces (IRS), a technology proven to improve VRU identification by small wavelength radar devices [68].

The last category of existing methodologies, indirect systems, leverages road infrastructure, such as sensors placed at intersections, to mitigate blind spots or signal blocking for AVs, enhancing communication between connected cars and VRUs [111]. Rippl et al. [249] proposed distinguishing pedestrians from bicyclists using features extracted from time-frequency analysis of radar sensor data under varying speed conditions. Additionally, Meissner et al. [203] assessed 3D measurements using a network of laser scanners to recognize pedestrians in real-time (measurement rate of 12.5 Hz). Their work employs a Gaussian mixture model and the Density Based Spatial Clustering of Applications with Noise (DBSCAN) algorithm [236] to cluster and identify VRUs.

Besides VRU detection, systems also exist to detect and notify jaywalking (i.e., pedestrians crossing undesignated areas). Using visible spectrum cameras and deep learning techniques, Mostafi et al. [214] proposed a multi-object and multi-camera tracking approach to identify jaywalkers and warn nearby connected vehicles, achieving 100% accuracy. Other studies focus on pedestrian safety within designated walkways with anomaly (e.g., skating and bicycles) detection methods [234, 255].

Additionally, some researchers focus on providing benchmarking datasets for VRU detection and classification. For example, Mammeri et al. [198] proposed a roadside perspective image dataset encompassing various less common VRU categories, including strollers and motorcycles. Figure 7 summarizes the most relevant studies in the literature, showing the datasets employed and the types of VRUs they focus on. For this, we have selected the fourteen most cited papers in these contexts, including works that deal with less common, but also vulnerable, traffic agents.

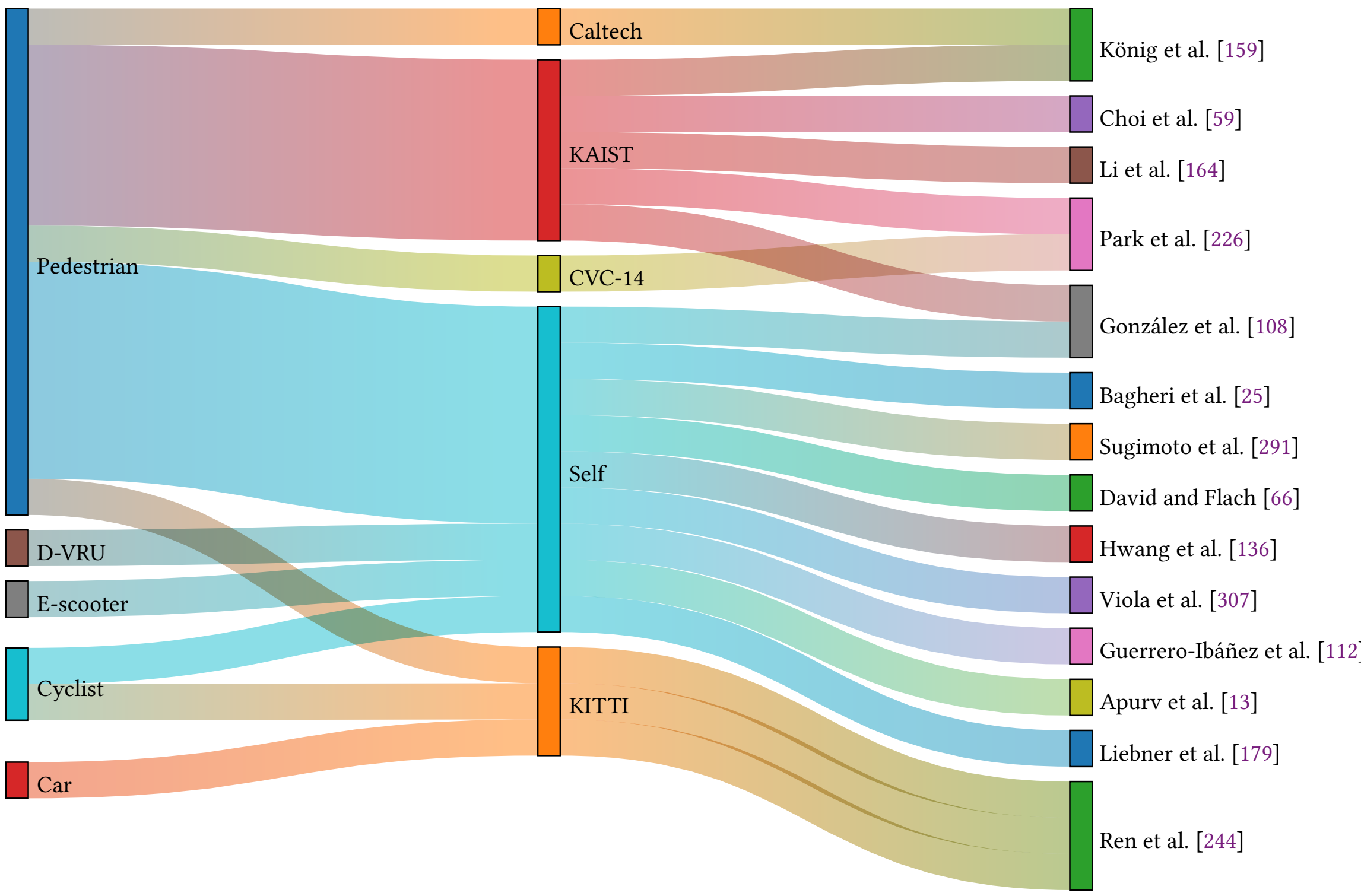

Fig. 7. The most relevant studies related to VRU detection and classification in the literature. "D–VRU" stands for disabled VRUs, while "Self" indicates private data collected by the author(s).

Despite various successful approaches using different types of sensors for VRU detection, the development of many camera-based datasets (see Table 9) has influenced the use of colored images in many traffic monitoring systems [283]. In this context, the perception of road agents is akin to an object detection task, where a target object is positioned in a scene and classified into a category [359]. One of the earliest pedestrian detection techniques was the frame-based Viola-Jones detector [307]. Other traditional algorithms identify VRUs using regions of interest (ROIs), built upon early object detection techniques such as background subtraction (BS) [230], HOG [64], and LBP [311].

With the rise of deep learning, artificial neural networks have replaced manual feature extraction in traditional machine learning algorithms and are now present in most proposed solutions [147, 157, 200]. We can divide these deep learning approaches into single- and two-stage detectors. Two-stage techniques, such as region-CNN (R-CNN) [102] and Faster R-CNN [245], first generate ROI from positive samples, followed by regional classification and location refinement [76]. The feasibility of using two-stage techniques for pedestrian detection has been investigated in some studies, including the work of Zhang et al. [352]. In this work, the authors observed a limited success of the Faster R-CNN method. They demonstrated that the region proposal network (RPN) — a backbone component of Faster R-CNN — together with boosted forests classifiers performed well for this detection task, achieving a satisfactory miss rate on the Caltech benchmark.

On the other hand, single-stage detectors identify ROI and extract their features within a single network, offering a more efficient, concise procedure [76, 244]. Notable algorithms in this category include the YOLO series [243] and single shot detector (SSD) [185], which can be used alongside multi-object tracking approaches such as simple online and realtime tracking (SORT) [34] and DeepSORT [316].

Although two-stage techniques allocate an entire stage for decision refinement, this does not necessarily result in more accurate bounding boxes detection compared to single-stage methods. For instance, Ren et al. [244] introduced the recurrent rolling convolution (RRC) architecture, a single-stage network that recurrently incorporates context into the bounding box regression process. When evaluated on the KITTI dataset, the RRC model outperformed SSD and ranked among the top methods in the KITTI detection challenge across multiple classes at the time of submission.

Another recent method that leverages context to improve its pedestrian detection is vision-language semantic self-supervision for pedestrian detection (VLPD), proposed by Liu et al. [181]. This approach integrates vision-language semantics (VLS), which employs a textual pseudo-label, obtained through cross-modal mapping, with an image data to acquire the existences of semantic classes in it. It also incorporates prototypical semantic contrastive (PSC), a prototype-based method that discriminates between pedestrians and other semantic classes, refining the inference process.

Meanwhile, other approaches emerged to enhance single-stage techniques. Liu et al. [186] introduced the asymptotic localization fitting network (ALFNet), which stacks a series of predictors to progressively refine anchor boxes. This process allows later predictors to be trained with higher intersection over union (IoU), leading to improved inference results and enabling ALFNet to surpass two-stage techniques while retaining the advantages of a single-stage approach. At the time of its publication, it achieved state-of-the-art results on the Caltech benchmark, outperforming all previous methods at an IoU threshold of 0.75.

In addition, some researches aim to improve single-stage detectors for specific scenarios or target classes, for example, a recent work developed by Liu and Shi [188] introduced VRU-YOLO, a redesigned architecture of YOLOv8 focused on small target feature extraction, motion blur, and shape variability for VRU detection. Their method demonstrated slightly better performance when compared to similar detectors and was also tested on specific remote sensing small target resources, such as VisDrone dataset [365].

Moreover, some studies employ multiple perception techniques, proposing sensor fusion strategies to identify VRUs. For example, Teixeira et al. [295] developed a hybrid (cloud and edge) architecture and algorithms to predict potential collisions between vehicles and VRUs. They combined input data from radar, LiDAR, and camera devices installed within AVs and the road infrastructure with positional information from VRUs' smart devices. As a result, their methodology was capable to achieve high speed (end-to-end communication under the 300 ms ceiling value) and scalability (stable results even with 1000 vehicles).

González et al. [108] demonstrated that combining visible (or RGB) and thermal (or infrared — IR) cameras improved pedestrian detection by decreasing average miss rate by 5% during the day. Similarly, other studies have evaluated the combination of illumination and radar sensors [21, 59, 164, 226] or provided multispectral-focused benchmarking resources like the KAIST dataset [136]. Specifically on this mentioned benchmarking, various visual-optical and thermal fusion methods were proposed. FusionRPN [159] employs multiple pretrained VGG-16 [284] CNNs to process three RGB channels and one IR channel as well as a decision-tree-based classifier to reduce false positives. Park et al. [226] employed a slightly different approach by fusing final detections from each spectral channel with channel weighting fusion (CWF) and accumulated probability fusion (APF). Furthermore, Zhang et al. [348] proposed a cyclic fuse-and-refine (CFR) module, designed to be integrated into any CNN architecture, incrementally improving the feature fusion process.

More recently, Zhang et al. [349] introduced the guided attentive feature fusion (GAFF) method, which employs inter- and intra-modality attention for feature fusion while maintaining a low computational cost, achieving a runtime of 9.34 ms on the KAIST dataset — significantly outperforming the CFR-based approach, which requires 50 ms. Similarly to the KAIST benchmark, Mirlach et al. [208] recently proposed the Roadside LiDAR-Visual-Thermal (R-LiViT) dataset, which extends previous work by incorporating LiDAR point clouds alongside RGB and IR images and is considered the first roadside dataset with this settings.

Techniques fusing data from LiDAR and camera are becoming more robust and accurate, being the state-of-the-art in the nuScenes dataset detection challenge [189, 314, 336]. LiDAR point-cloud data does not capture information such as colors and textures, which is useful data for classifying proposals. Meanwhile, cameras capture these types of information, but they fail to capture depth information, which LiDAR can obtain more efficiently [99]. Therefore, combining data from these different types of sensors to detect objects can reduce the impact of deficiencies in individual sensors [334]. TransFusion [26], one of the first methods fusing camera and LiDAR data to obtain competitive results on nuScenes, firstly predicts 3D bounding boxes from the point cloud and extracts image deep features, then enhances the bounding boxes predictions using transformers and the image features.

Another foundational sensor fusion method is BEVFusion [178, 189]. In this approach, data from both sensors are fused to generate a shared bird's-eye view (BEV) representation that preserves semantic and geometric information — often lost in traditional LiDAR-to-camera and camera-to-LiDAR transformations. Additionally, BEVFusion can be applied to other tasks (e.g., map segmentation) and has been considered a baseline method for several other camera and LiDAR fusion strategies, such as edge-aware lift-splat-shot (EA-LSS) [125], FusionFormer [124], MV2DFusion [314], IS-Fusion [336], and SparseLIF [350].

In EA-LSS [125], two modules are proposed to improve the depth prediction of 3D bounding boxes. These modules are responsible for reducing the difficulty of interference in regions of high-depth changes. In BEVFusion4D [44], point-cloud BEV features are used as a guide for generating camera BEV features, improving its spatial information. Another approach is presented in FusionFormer [124], which uses transformers to fuse point cloud and camera data into BEV format more efficiently. Both BEVFusion4D and FusionFormer use a temporal fusion encoding module to merge previous extracted BEV features to enhance the final bounding boxes.

There are other possible data fusion model approaches that have achieved performance advances. One is IS-Fusion [336], which extracts objects specific features, merging them with the scene point cloud and cameras features to predict more accurate bounding boxes. Another approach is SparseLIF [350], which uses a sparse network for training end-to-end, having a module to evaluate uncertainty in predictions from different sensors, contributing to the filtering of noisy data. In MV2DFusion [314], to avoid bias toward one modality, a 2D object detector generates bounding boxes for image data and a 3D object detector for the point cloud data. These detections are then input to a fusion module, which generates the final 3D bounding boxes.

Continuing on cross-modal fusion studies, [175] introduced DeepFusion, a deep feature-level fusion method to combine camera and LiDAR data in the Waymo dataset. However, differently from previous approaches, DeepFusion merges features extracted from point-cloud data into camera features rather than raw 3D points. Additionally, their study offers novel techniques to enhance spatial alignment between LiDAR and camera spaces, as well as feature correspondence across modalities using cross-attention.

Another recent method is local-to-global fusion network (LoGoNet), proposed by [172]. This work introduced the possibility of producing fine-grained features from specific portions of the images, potentially improving the discrimination of small and occluded objects. This is achieved through a local fusion (LoF) process, where detection proposals are subdivided into uniform grids and their centers are projected onto the image plane. As well as in DeepFusion, LoGoNet also employs cross-attention mechanism to selectively integrate the most relevant visual and LiDAR features.

Some recent studies use fusion techniques with 4D radar data. The MSSF [180] model created a new camera and 4D radar fusion block, which can effectively merge these different types of data, surpassing the classical LiDAR detection methods. Another method in this scope is L4DR [130], which fused 4D radar data with LiDAR point clouds, enhancing the detection accuracy in adverse weather conditions.

We can group the majority of data fusion techniques into (i) pixel-level, (ii) early fusion, (iii) halfway fusion, and (iv) late fusion [5]. Pixel-level fusion is the only method not typically employed with deep learning algorithms, though some researchers still use it [94]. Early (or feature-level) fusion combines sensor inputs into a single network. Halfway (or middle-level) fusion feeds data from multiple sensors separately into the network, combining them at an intermediate layer. In late (or decision-level) fusion, sub-networks process each sensor input, and only the output layers are combined for classification. Table 14 summarizes these data fusion strategies, including their advantages and disadvantages.

Table 14. Main data fusion techniques for VRU detection and classification.

| Method | Pros | Cons |
|---|---|---|
| Pixel-level | Enables distinguishing features impossible to perceive with any individual sensor [297]. | Usually not applicable with deep learning algorithms [5];<br>Requires preprocessing steps such as Image Registration [297]. |
| Early | Does not use sub-networks;<br>Suitable for sparse and depth images [98]. | The output of different sensors does not have the same size and each sensor has its own properties [229]. |
| Halfway | Does not use sub-networks. | Uncertainty about optimal intermediate fusion point. |
| Late | Results are considered more trustworthy [283]. | Use of several classifiers [83]. |

To assess the detection and classification of VRUs and other traffic agents, systems developed for this purpose can utilize specific evaluation metrics, such as those defined by Caesar et al. [43]:

- **Intersection over union (IoU)**: a ratio between the overlap area and the union area considering the ground-truth and predicted bounding boxes;
- **Average translation error (ATE)**: euclidean center distance 2D in meters.
- **Average scale error (ASE)**: calculated as $1 - IoU$ after aligning centers and orientation;
- **Average orientation error (AOE)**: smallest yaw angle difference between prediction and ground-truth in radians;
- **Average velocity error (AVE)**: absolute velocity error in meters per second;
- **Average attribute error (AAE)**: calculated as $1 - accuracy$;
- **Average precision (AP)**: average precision of a model across different levels of recall;
- **Mean average precision (mAP)**: average AP metric for different classes and IoU thresholds; and
- **nuScenes detection score (NDS)**: metric employed in nuScenes detection challenge. It is calculated as a weighted sum of the mAP metric and the ATE, ASE, AOE, AVE, and AAE error metrics.

Table 15. Performance comparison of VRUs detection models. The abbreviation “AMR” stands for average miss rate. VRUs “P”, “C”, and “M” abbreviations stands for pedestrian, cyclist, and motorcyclist, respectively. LC Fusion is “LiDAR+Camera fusion” and MS Fusion is “Multi-spectral fusion”

| Dataset | VRUs | Model | Backbone | Ped. AP | Cyc. AP | NDS | mAP | AMR |
|---|---|---|---|---|---|---|---|---|
| KITTI | P,C | LoGoNet [172] | CNN, Attention, RCNN, LC Fusion | 64.55% | 84.00% | | | |
| | | RRC [244] | CNN | 75.33% | 76.47% | | | |
| nuScenes | P, C, M | SparseLIF [350] | CNN, Attention, FPN, LC Fusion | | | 77.0% | 74.4% | |
| | | MV2DFusion [314] | CNN, Attention, FPN, LC Fusion | | | 76.7% | 74.5% | |
| | | IS-Fusion [336] | CNN, Transformer, LC Fusion | | | 75.2% | 73.0% | |
| | | FusionFormer [124] | CNN, Transformer, LC Fusion | | | 75.1% | 72.6% | |
| | | BEVFusion-4D [44] | CNN, FPN, Transformer, LC Fusion | | | 74.7% | 73.3% | |
| | | EA-LSS [125] | CNN, Attention, LC Fusion | | | 74.4% | 72.2% | |
| | | BEVFusion [189] | CNN, FPN, LC Fusion | | | 72.9% | 70.2% | |
| | | BEVFusion [178] | CNN, FPN, LC Fusion | | | 71.8% | 69.2% | |
| | | TransFusion [26] | CNN, Transformer, LC Fusion | | | 71.7% | 68.9% | |
| Waymo | P, C | LoGoNet [172] | CNN, Attention, RCNN, LC Fusion | 81.55% | 73.89% | | | |
| | | BEVFusion [189] | CNN, FPN, LC Fusion | 79.06% | 76.00% | | | |
| | | DeepFusion [175] | CNN, Attention, LC Fusion | 79.16% | 75.47% | | | |
| KAIST | P, C | GAFF [349] | CNN, Attention, MS Fusion | | | | | 10.62% |
| | | CFR [348] | CNN, MS Fusion | | | | | 10.05% |
| | | CWF+APF [226] | CNN, MS Fusion | | | | | 31.36% |
| | | FusionRPN+BDT [159] | CNN, MS Fusion, Boosted Decision Trees | | | | | 29.83% |
| Caltech | P | VLPD [181] | CNN, Self-supervision, Text encoder | | | | | 2.3% |
| | | ALFNet [186] | CNN | | | | | 6.1% |
| | | RPN+BF [352] | CNN, Boosted Forest | | | | | 23.5% |

A recent and emerging research focus in the autonomous driving field involves the application of large language models (LLMs), a generative AI technique, in traffic scenarios. This approach aims to shift from a purely data-driven paradigm to one that is knowledge-based and more generalized [170]. LLMs offer notable real-world understanding and logical reasoning, which make them highly suitable for self-adaptive systems (SaS) [169], including those utilized in AVs.

Additionally, multi-modality can be merged with LLMs to develop multimodal large language models (MLLMs), which can be employed for various tasks, including object detection and tracking [71, 320], semantic understanding [215], multitasking and end-to-end learning [48, 138], planning, evaluation, and more [61, 333, 362].

In VRU detection and classification, Ding et al. [71] introduced a high-resolution understanding method in MLLMs for autonomous driving (HiLM-D). This approach aims to assess risk object localization and intention and suggestion prediction (ROLISP) from image data, a task that can encompass VRUs, such as cyclists. The evaluation was performed using a custom dataset generated by the authors, testing objects of various sizes. The proposed method achieved a mean IoU of 59.2. Furthermore, Kim et al. [148] proposed a language-driven framework for pedestrian detection from multispectral data. By the use of the MLLM GPT4-V[9] and a chain-of-thought prompting strategy, the authors were able to improve the precision of detections in standard multispectral datasets.

Finally, Table 15 indicates the performance on specific datasets reported by the main VRUs detection methods previously mentioned in this section. Furthermore, the table also suggests each work's focused VRU type and backbone methods.

### 6.1 Final considerations

The safety of VRUs in connected environments is becoming increasingly dependent on the evolution of AV technologies and communication paradigms. Deep learning models, sophisticated hardware, and multimodal systems are necessary to detect and classify VRUs accurately. However, challenges such as network congestion, energy efficiency, real-time inference, and context information exchange must be addressed by future studies to guarantee the safety of this group of users in increasingly complex traffic environments. Although approaches such as integrating LLMs and MLLMs are considered unconventional and have shown transformative potential, they are still in the early stages of development and require further investigation aiming to overcome several limitations related to privacy, scalability, hallucination, latency, or computational complexity [74, 196].

## 7 VRU ACTION, BEHAVIOR, AND INTENTION PREDICTION

In addition to detecting the presence of VRUs in the traffic environment, AVs must predict potential future harms for nearby users caused by their actions or inactions [82]. To this end, action, behavior, and intention prediction techniques are employed to improve safety in mixed road scenarios. According to Sharma et al. [279], "action" refers to identifying physical movements (e.g., waving hands), "behavior" denotes observable events in response to stimuli, and "intention" reflects an intrinsic user's state of mind.

By anticipating possible decisions of VRUs (e.g., crossing movements) and other vehicles (e.g., lane changes), AVs may have sufficient time to plan appropriate maneuvers [82]. However, achieving this is complex. Beyond the limitations of each type of sensor employed by AVs, the main challenge lies in perceiving cues in typical traffic contexts to avoid severe collisions [235].

[9]Developed by OpenAI. Available at https://openai.com/ (accessed on June 24, 2025).

Several studies aim to explain human conduct on the road, focusing on different types of road users, such as pedestrians [32] and drivers [118]. These studies employ various methods, including questionnaires and other approaches to collect self-reported data from participants [121], observing actions in natural scenarios [285], and utilizing computational technologies such as deep learning [344] and inverse reinforcement learning (IRL) [219].

Some works aim to model the dynamics of traffic accidents to identify high-risk crash areas within cities (e.g., city centers and public spaces) and to analyze how these risks may vary based on the day of the week and time of day [114]. Generally, road structure [197], user interactions [81], and social, cultural, or demographic factors [241] often influence VRUs' decisions and can not always be captured by AVs. Additionally, some studies provide guidelines for designing future AVs from VRUs' perspectives [132, 210].

We can use various types of information as input for VRU behavior prediction models. We categorize these input types into six groups, similarly to Ridel et al. [248]: "dynamics", "body", "pose", "environment", "social-related", "head orientation", and "gesture" (Table 16). Dynamics and body-related (e.g., head, gestures) data can be combined to model VRU awareness. Moreover, incorporating environmental aspects and interactions between surrounding road agents can generate robust solutions, known as dynamics, awareness, and scene understanding approaches.

Table 16. Description of different types of inputs for VRU behavior prediction.

| Input type | Description |
|---|---|
| **Dynamics** | The VRU dynamics can indicate its trajectory and intention. Some studies successfully predicted VRU behavior by using multiple consecutive image frames and evaluating VRUs' position in each. Additionally, we can use VRU speed and acceleration estimation for intention and trajectory prediction [248]. |
| **Body pose** | The body pose of humans is essential in many computer vision and can be used for human action recognition, human tracking, human-computer interaction, gaming, sign languages, and video surveillance [216]. Human pose estimation consists of localizing body key points and identifying the posture of people [216]. This information can enhance the prediction of VRUs' intentions and trajectories [274, 310]. |
| **Environment** | The environment in which VRUs and vehicles are inserted can influence the interactions between them. For instance, traffic signals, bicycle infrastructure, and parked cars may influence their interactions [60]. |
| **Social related** | Some studies state that social interactions may influence the VRUs' decisions. For example, these interactions can include pedestrians trying to be far from others, avoiding others coming towards them, or following the flow of other pedestrians [248]. |
| **Head orientation** | The head orientation consists of classifying the direction in which a person is looking. Saleh et al. [257], for example, used the classes "front", "back", "left" and "right" to identify the head orientation of VRUs. However, the head direction can only sometimes predict the user's intention because the user can look at an advertisement or search for someone. |
| **Gesture** | VRUs can use gestures to communicate their intentions to drivers of nearby vehicles. For example, cyclists can use hand signals to indicate whether they will stop, go left, or go right [112]. |

Many studies combine different input types to achieve better results [80]. To do this, these diverse pieces of information must undergo a fusion process before being used in behavior prediction models. A common method in the literature is "concat fusion" [80], which merges different types of inputs without considering the relevance of each. For example, Rasouli et al. [239] improved an intention prediction model's accuracy by concatenating body pose, ego-vehicle speed, and environmental inputs.

Another well-known method for combining different input types is "attentive fusion", which assigns greater relevance to certain types of inputs, leading to improvements in more recent works [80]. This approach is employed by Zhou et al. [363], which used a transformer architecture for intention prediction, fusing ego-vehicle, pedestrian, and environment inputs.

As stated by Korbmacher and Tordeux [153], early studies on VRU intention prediction, particularly for pedestrians, relied on direct observations and photographs to enhance understanding of their behavior. Subsequently, simulation models such as force-based [120], queuing [191], and transition matrix [95] were developed, categorizing typical prediction techniques into (i) macroscopic, (ii) mesoscopic, and (iii) microscopic. The first two groups analyze aggregated levels, while the latter focuses on individual VRU motion and can be further divided into acceleration-based [144], velocity-based [228], and decision-based models [39].

Traditional behavior analysis methods include Bayesian filters [266], hidden Markov models (HMM) [366], latent-dynamic conditional random fields (LDCRF) [269], and Gaussian processes and their variations [235, 266]. These methods were employed through (i) physics-based (or dynamical) and (ii) goal-driven (or planning-based) models. Physics-based approaches require precise modeling and do not perform well on long-term predictions [279]. Conversely, the final destination of the VRU — a challenging variable to infer for moving vehicles [5] — must be known in planning-based approaches.

Furthermore, deep learning methods have been proposed for VRU behavior anticipation, leading to data-driven approaches capable of achieving high performance even in unmodeled scenarios. As noted by Sharma et al. [279], proposed solutions include CNNs [82, 304], LSTM [6, 28, 240, 259], game-theory-based models [193], LSTMs with attention mechanisms [85, 87], autoencoders [161, 168], graph neural networks (GNN) [8, 305, 361], generative adversarial network (GAN) [115, 256], and transformer-based models [103].

Figure 8 summarizes the most relevant studies in VRU action, behavior, and intention prediction, including their main tasks and datasets used. For this, we have selected the twenty most cited papers in these contexts. In the following sections, we analyze the most relevant studies on VRU behavior prediction, categorized into three different time frames, as defined by Zhang and Berger [345].

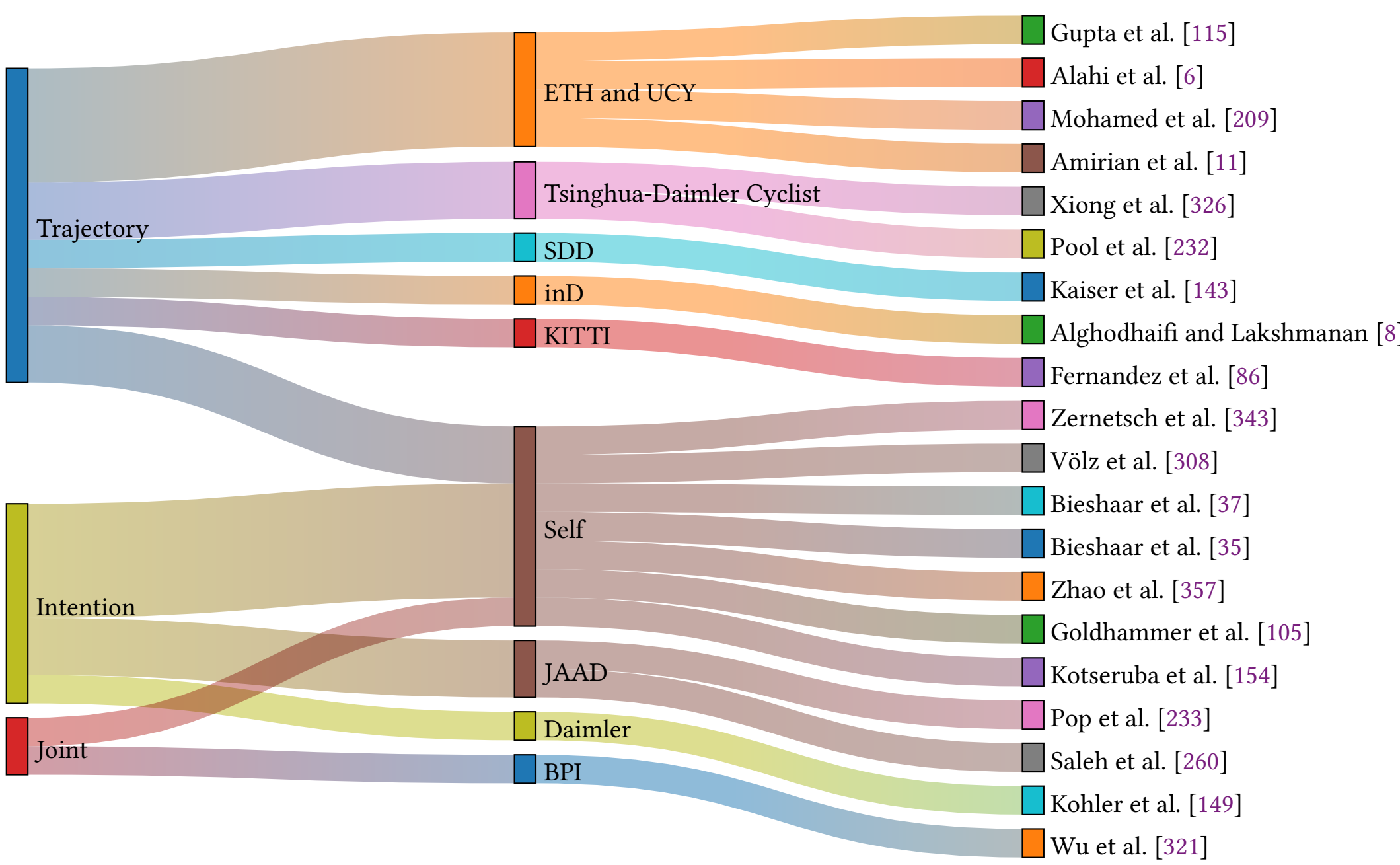

Fig. 8. The most relevant studies related to VRU action, behavior, and intention prediction in the literature.

### 7.1 Intention prediction

The task of VRU intention prediction can be considered a classification problem [80, 345]. However, some studies approach it as a combination of classification and trajectory prediction. Most research aims to classify whether an identified VRU intends to cross the road. Some studies introduce intermediary classes, such as "starting to cross the road" [35, 37, 105, 149, 233] and "ambiguous intention" [233], though the primary focus remains on predicting crossing or not crossing.

Certain studies employ human gestures for VRU intention classification. By detecting VRUs and extracting hand information, we can identify whether a gesture is made and what intention it indicates. For example, Ashtekar et al. [20] created a dataset with videos of a person riding a bicycle and gesturing with various intentions. They used a CNN for gesture prediction, classifying gestures into "stop", "give way", "left", "right", "road hazards", and "slow down" categories. Similarly, Guerrero-Ibáñez et al. [112] presented a model to identify disabled VRUs (D-VRUs) and their intentions, classifying them as "stop", "I want to cross", "you cross", and "I will cross first". They used a LSTM and the MediaPipe framework.

Some early works merged the task of detection and intention prediction, using bounding boxes containing classes of VRU intention, such as classes for pedestrians crossing or not crossing the street. In this way, these algorithms can be evaluated through detection analyses, such as using the AP metric. An example of this approach can be seen in [233], which adapted the RetinaNet model for object detection and intent classification. Another possibility was explored by Saleh et al. [260], who used the YOLO model to detect VRUs and an adaptation of DenseNet to classify bounding boxes.

With the creation of the action prediction benchmark [155], many studies started using standard metrics in the classification task to evaluate intention prediction methods. These metrics include the number of true positives (TP) (i.e., correctly predicting crossing the street) and true negatives (TN) (i.e., correctly predicting not crossing the street). Additionally, they compute the number of false positives (FP) and false negatives (FN), which are the incorrect classifications of positive and negative cases. The most frequently used metrics are accuracy, precision, recall, and F1-score.

Several recent works successfully used attention mechanisms to obtain state-of-the-art results [10, 155, 332]. Attention excels by selectively focusing on relevant information, enhancing model performance. Another recent approach is based on LLMs, which, in addition to VRU detection, can also be applied to predicting the intentions of VRUs. As Huang et al. [127] reported, the vision-language model (VLM) GPT-4V was tested on JAAD dataset and achieved satisfactory results for diverse actions.

Finally, Table 17 indicates the performance on datasets reported by the main intention prediction methods previously mentioned in this section. Furthermore, the table also suggests each work's focused VRU type and backbone methods.

### 7.2 Trajectory prediction

Trajectory (or motion) prediction involves computing a detailed spatiotemporal representation of VRU behavior, which is essential across various research disciplines [345]. This task is often described as a sequence of intention predictions [54] and is typically associated with long-term analysis scenarios. According to Zhang and Berger [345], we can categorize path forecasting approaches as (i) regression, (ii) classification, or (iii) discrete variable modeling.

When modeled in a regression context, we can represent the system's output as pairs of position coordinates [103, 221, 258, 259]. Although this approach is simple, it struggles to capture spontaneous behaviors of targets. Alternative approaches include projecting outputs using unimodal [6, 209, 305, 346] or multimodal [11, 115, 356] statistical distributions. The latter overcomes the poor generalization ability of the former despite requiring more computational power.

Table 17. Performance comparison of VRUs intention prediction models. *: Average Precision of Pedestrian is Crossing bounding boxes

| Dataset | VRUs | Model | Backbone | Acc | AUC | F1 | AP* |
|---|---|---|---|---|---|---|---|
| JAAD | Pedestrian | $PPCI_{att}$ [10] | LSTM, Attention | 0.81 | 0.78 | 0.75 | |
| | | GPT-4V [127] | LLM, Attention, GPT | 0.57 | 0.61 | 0.65 | |
| | | Ahmed et al. [4] | LSTM | 0.89 | | | |
| | | MTL [267] | RNN | 0.90 | 0.95 | 0.76 | |
| | | Yang et al. [332] | GRU, CNN, Attention | 0.83 | 0.82 | 0.63 | |
| | | PCPA [155] | RNN, Attention, CNN | 0.85 | 0.86 | 0.68 | |
| | | ST-DenseNet [260] | YOLO, CNN | | | | 73.78% |
| | | Pop et al. [233] | CNN, LSTM | | | | 65.57% |
| PIE | Pedestrian | $PPCI_{att}$ [10] | LSTM, Attention | 0.91 | 0.89 | 0.84 | |
| | | Ahmed et al. [4] | LSTM | 0.91 | | | |
| | | $CIPTLU_{mt}$ [301] | LSTM | 0.90 | 0.87 | 0.82 | |
| | | MTL [267] | RNN | 0.91 | 0.93 | 0.82 | |
| | | TED [3] | Attention | 0.91 | 0.91 | 0.83 | |
| | | Yang et al. [332] | GRU, CNN, Attention | 0.89 | 0.86 | 0.80 | |
| | | PCPA [155] | RNN, Attention, CNN | 0.87 | 0.86 | 0.77 | |

Another possibility is to encode trajectory forecasting as a high-dimensional discrete variable using grid-based representations or by transforming VRUs' velocity into bins [199, 288]. The motion prediction task can also be considered a labeling problem, with input represented through one-hot vectors. However, classification models generally perform worse than regression models [103].

Various criteria can group deep learning-based strategies for trajectory prediction. Rudenko et al. [254] divided these studies into (i) sequential (or time-series) and (ii) non-sequential approaches. The first group assumes that the current state of the target is based on a series of chronological states and can learn motion patterns in specific environments (i.e., local transition patterns) or general spaces (i.e., location-independent behavioral patterns). Non-sequential approaches, on the other hand, capture distributions of motion over long-term, complete trajectory data.

Bighashdel and Dubbelman [38] categorized path forecasting strategies that incorporate data-driven methods into (i) interaction-based, (ii) path-planning-based, and (iii) intention-based models. Interaction-based models consider the interactions between VRUs and their environment as the main influencing factors for their behavior, including solutions such as behavior-CNN [335], social-grid-LSTM [56], and context-aware social-LSTM [28]. Path-planning-based methods assign VRUs' behavior based on their final destination [129, 369]. Intention-based models focus on predicting the following VRU intentions to form a sequence of movements [308].

Although most deep learning-based path forecasting methods output deterministic trajectories, some compute probabilistic estimations. As reported by Golchoubian et al. [104], predictions using distribution functions can include uncertainties related to pedestrians' trajectories. In other words, the networks output parameters to define the results for a given distribution, such as bi-variate Gaussian [57, 204] or Cauchy [290] distributions.

Furthermore, we can predict VRUs' trajectories using confidence regions, as presented by Schneegans et al. [265], who employed quantile surface neural networks (QSN) to forecast cyclists' trajectories and plan AV lane movements. Another example is introduced by Zernetsch et al. [343], where a neural network outputs a numerical quantification of uncertainty for predictions, later compared to a normal statistical distribution to analyze its reliability.

As mentioned before, behavior anticipation models can encode features from many sources. Among the path forecasting methods, Goldhammer et al. [106] employed polynomial least-squares approximation from camera-based head tracking data to predict pedestrian location up to 2.5 seconds ahead. Czech et al. [63] proposed the behavior-aware pedestrian trajectory prediction (BA-PTP), an approach based on a person's head orientation, body orientation, and pose, outperforming earlier state-of-the-art methods on the PIE dataset.

Head and body data can be applied with smart devices to enhance predictions, as demonstrated by Bieshaar et al. [36]. In their study, head orientation history captured from surveillance cameras and positional data from smartphones formed a cooperative system that achieved lower delay and higher F1-score.

Other factors, such as demographics (e.g., age and gender) and social characteristics, were considered by a data-driven approach proposed by Chen et al. [52], which employs attention mechanisms to assign weights between input features automatically. Pool et al. [232], on the other hand, considered road topology data to enhance the accuracy of different probabilistic traditional methods for cyclists' path estimation.

The main evaluation metrics employed in VRU trajectory prediction are mostly based on distance or geometric comparison to a ground-truth (i.e., real) movement. As detailed by Sharma et al. [279], Zhang and Berger [345] and Schuetz and Flohr [268], studies can use — but are not limited to — the following indicators:

- **Average displacement error (ADE)**: also known as mean squared error (MSE), computes the distance between ground-truth and prediction trajectories for each predicted time step;
- **minADE$_k$**: an application of the original ADE to multimodal scenarios, in which only the first $k$ predictions with the lowest Euclidean distance are considered;
- **Final displacement error (FDE)**: considers only the ADE at the last estimated time step;
- **minFDE$_k$**: similarly to minADE$_k$, it considers only the top $k$ closest predictions but only at the final time step;
- **Center mean square error (C$_{MSE}$)**: calculates the MSE from the ground-truth path considering the center of the target's bounding box during the entire prediction duration;
- **Center final mean squared error (CF$_{MSE}$)**: considers only the C$_{MSE}$ at the last estimated time step;
- **Miss rate (MR)**: a ratio of predictions in which the FDE exceeds a threshold, such as 2 meters [364]. This metric can also be decomposed into longitudinal or latitudinal thresholds, and for $k$ different trajectories (MR$_k$) in multimodal problems; and
- Specific evaluation metrics for multimodal contexts, such as coverage and Gaussian-based assessments, as detailed by Huang et al. [128].

Some studies use multiple object tracking (MOT) metrics [1, 281], such as:

- **Identity switches (IDS)** [173]: the number of times that a tracked trajectory changes its matched ground-truth identity;
- **Multiple object tracking accuracy (MOTA)**: quantify the accuracy of object tracking;
- **Multiple object tracking precision (MOTP)**: a measure of the precision of object tracking;
- **Higher order tracking accuracy (HOTA)** [192]: geometric mean of detection accuracy and association accuracy, averaged across localization thresholds;
- **ID F1-score (IDF1)** [251]: the ratio of correctly identified detections over the average number of ground-truth and computed detections;
- **Mostly tracked targets (MT)**: ratio of ground-truth trajectories that are covered by a track hypothesis for at least 80% of their respective life span;
- **Mostly lost targets (ML)**: ratio of ground-truth trajectories that are covered by a track hypothesis for at most 20% of their respective life span;
- **False alarms per frame (FAF)**: average number of false alarms per frame; and
- **Track fragmentation (Frag)**: total number of times a trajectory is interrupted.

A novel study by Korbmacher et al. [152] demonstrated that deep learning-based methods applied with distance metrics might not be suitable for high-density pedestrian scenarios (i.e., environments with a significant presence of individuals and low degree of freedom). They proposed a continuous metric based on time-to-collision between two pedestrians. This new approach addresses limitations of previous metrics for pedestrian trajectory analysis, such as the inability to differentiate severity between collisions and to detect scenarios in which a prediction causes multiple crashes.

As outlined in the previous sections, applications of LLMs and MLLMs are emerging in the context of perception for urban environments. This includes the possibility of predicting forthcoming paths for VRUs. For instance, Wu et al. [320] introduced an object-centric language prompt approach to forecasting object trajectories. In line with this, they also proposed NuPrompt, an extension of the nuScenes dataset enhanced with language descriptions. Their method achieved performance metrics of up to 0.127 AMOTA, 1.361 AMOPT, a recall rate of 43.5%, and 0.135 MOTA.

Similarly, Chib and Singh [58] used a LLM to incorporate motion cues for pedestrian trajectory prediction. Other recent works further expand this direction: SPU-BERT [218] leverages socio-physical understanding through a dual-BERT architecture to provide fast and accurate multi-trajectory prediction, while LMTraj [22] reformulates trajectory forecasting as a language-based question-answering task, demonstrating competitive results via prompt-driven inference over multimodal inputs.

Finally, Table 18 indicates the performance on specific datasets reported by the main trajectory forecasting methods previously mentioned in this section. Furthermore, the table also suggests each work's focused VRU type and backbone methods.

### 7.3 Joint prediction

Joint or multi-task prediction leverages both intention and trajectory predictions to enhance the accuracy beyond what is achievable by either method alone. We can categorize this concept into two primary frameworks [345]. One approach involves using the same features to simultaneously predict the trajectory and label the intention within a single network, potentially reducing computational costs. The other approach consists of separately predicting intention and trajectory, then using each to refine and improve the other.

Several studies have explored these approaches. For instance, Liang et al. [177] proposed a model called Next, which predicts trajectory and actions simultaneously using a network enriched with visual information features. This model offers multiple benefits, including better overall path prediction and the ability to predict future actions.

In contrast, Goldhammer et al. [105] employed different multilayer perceptron (MLP) networks to predict the current motion state and trajectory of VRUs, and then combined the results to generate a final trajectory prediction. This highly modular approach replaces intention or trajectory prediction models with others that produce better results while maintaining the same general prediction operation. Although this method did not significantly improve prediction quality compared to other approaches, the authors highlighted the potential of joint prediction. They highlighted the benefits of modularized predictions and the integration of diverse data sources when fusing the predictions.

Other studies have also segregated intention and path estimation into different branches. Wu et al. [321] and Kotseruba et al. [154] focus on separate yet complementary prediction tasks. Despite the potential for joint prediction systems to outperform single-task methods, computational complexity remains a concern. Treating intention and trajectory predictions as separate and complementary tasks requires multiple networks and extensive preprocessing and data integration procedures. This can lead to slow training times, making such approaches less suitable for some scenarios.

### 7.4 Final considerations

Despite recent advances, the literature still needs to fill many gaps related to VRU detection and classification, as well as action, behavior, and intention prediction. For intention estimation methods, only a few studies (e.g., [20, 112]) evaluate contextual cues such as hand or head gestures between pedestrians and drivers, which are informal communications that can convey intentions [112]. Furthermore, the absence of a well-defined set of possible intentions and their measurements (e.g., crossing intention) can make behavior anticipation vague, as it often predicts only single actions [54].

Moreover, many studies use different datasets, which in many cases are private, as illustrated in Figure 8. Consequently, many algorithms are trained only in specific scenarios and may generalize poorly to others. Emphasizing this issue, Gesnouin et al. [97] show that state-of-the-art intention prediction models perform worse when evaluated on datasets different from those used for training. However, these models should be universal, working in various scenarios with different road structures, traffic signals, and other conditions [241].

Table 18. Performance comparison of VRUs trajectory prediction models. The Abbreviations “m” and “px” refer to meters and pixels, respectively. VRUs “P” and “C” abbreviations stands for pedestrian and cyclist, respectively.

| Dataset | VRUs | Model | Backbone | ADE | FDE |
|---|---|---|---|---|---|
| ETH | P | LG-Traj [58] | LLM, Gaussian mixtures | 0.38m | 0.56m |
| | | Social-STGCNN [209] | Spatio-temporal graph and time-extrapolator CNNs | 0.64m | 1.11m |
| | | Social-Ways [11] | LSTM, GAN, attention pooling | 0.39m | 0.64m |
| | | Sophie [256] | LSTM, GAN, attention mechanism | 0.70m | 1.43m |
| | | Social-GAN [115] | LSTM, GAN | 0.60m | 0.87m |
| | | Social-LSTM [6] | LSTM | 0.50m | 1.07m |
| SDD | P | LG-Traj [58] | LLM, Gaussian mixtures | 7.80px | 12.79px |
| | | Sophie [256] | LSTM, GAN, attention mechanism | 16.27px | 29.38px |
| | | Social-GAN [115] | LSTM, GAN | 27.246px | 41.440px |
| | | DESIRE [161] | CNN, RNN, conditional variational auto-encoder | 19.25px | 34.05px |
| | | Social-LSTM [6] | LSTM | 31.19px | 56.97px |
| KITTI | P, C | Fernandez et al. [86] | CNN, LSTM, BEV mapping | 0.525m | 1.145m |
| inD | P | HSTGA [8] | LSTM, temporal-LSTM, graph attention | 0.23m | 0.29m |
| Tsinghua-Daimler Cyclist | P, C | Xiong et al. [326] | RNN, LSTM | 1.22px | |

Similarly, strategies developed to anticipate VRU trajectories must still address significant gaps. Understanding pedestrian variability is crucial for safety improvements since there is still a discrepancy between their expected and actual behavior [242]. The ability to handle abrupt motion changes, noise features from detection systems, and varying target densities (i.e., crowds and single individuals) remains challenging. Furthermore, incorporating visual (or appearance) behaviors could strengthen predictions for scenarios without past trajectories (e.g., stationary users).

Another pressing issue is occlusion, which limits the visibility of VRUs and significantly degrades detection performance, especially for pedestrians who are more than 50% occluded, leading to an increased false negatives rate [101]. Considering that many motion prediction methods rely on object detection as a first required task for their inference, occlusion may introduce uncertainty into trajectory forecasting. Recent work has explored occlusion-aware motion planning using phantom agent modeling and sequential reasoning [212], offering promising tools to proactively assess risk in partially observable areas. These advances may enhance trajectory-based safety mechanisms and indirectly support downstream components such as active braking control [122].

In conclusion, while joint prediction systems hold promise for improving the accuracy and robustness of VRU action, behavior, and intention predictions, careful consideration of computational resources and system design is imperative. Balancing the benefits of enhanced prediction accuracy with the practical constraints of computational efficiency will be essential for successfully deploying these systems in real-world applications.

## 8 CONCLUSION

Ensuring the safety of VRUs is essential for adapting densely populated and increasingly congested urban environments with strategies and technologies that protect the most vulnerable, minimize accidents, and save lives. In this context, this paper has presented a comprehensive bibliographical survey, highlighting crucial points for enhancing VRU security. We have demonstrated that the communication ecosystem between vehicles and VRUs has developed promisingly, leading to safer and more integrated systems that enable harmonious interactions.

Our analysis of sensor types and datasets reveals multiple methods for collecting data related to the road ecosystem, each with distinct advantages and disadvantages suited to various scenarios. The available datasets incorporate critical aspects of VRU research, varying in sensor types, and target objects. Furthermore, studies in the literature show variations between real and synthetic data, location information, and viewpoints. These differences arise from the sensor's installation location, which can be fixed (e.g., poles) or dynamic (e.g., on vehicles or drones), offering diverse data collection perspectives from horizontal and aerial views. Notably, most VRU research predominantly uses cameras and LiDAR as primary sensors.

In addition to real data collection, we explored the generation of synthetic data and scenarios using various simulation environments documented in the literature. These tools enable the analysis of traffic user behaviors and anticipating results before implementing sensors and infrastructure in real-world settings. Simulation environments can also combine real and simulated information, broadening the strategies to enhance VRU safety. Our survey identified CARLA and SUMO as the most frequently used simulation environments.

Integrating data from diverse sources, including literature datasets and real-time traffic environments, is essential for tasks related to traffic perception, such as detection, tracking, classification, and intention prediction. Our study analyzed existing research, providing insights into the influencing factors and employed methodologies. These insights help anticipate the behaviors and future actions of traffic participants while addressing challenges such as varying lighting conditions, climate changes, and obstacles. It was observed that most of the recent works in this area employs deep neural networks.

We recommend that future studies focus on seamless integration for interaction between VRUs and vehicles. Additionally, sensor fusion can leverage the strengths of various sensors, enhancing vehicle perception. It is important to note that most studies and datasets are collected in countries with organized traffic systems [182]. Thus, expanding research to countries with chaotic traffic and cultural behaviors that increase accident risk, such as Brazil and India, is necessary. For instance, in Brazil, motorcycles often travel between vehicles, and pedestrians frequently cross streets outside designated crossings. Similarly, in India, vehicles, especially motorcycles and auto-rickshaws, commonly navigate between lanes and are often overloaded.

Another critical point is that many studies and proposed datasets focus on VRU safety from the perspective of AVs or those equipped with extensive sensor technology. However, many vehicles, particularly those popular among lower-income groups, lack sensors, especially in several countries. This scenario may take years to change. Our research indicated that while some studies explore roadside units, they remain in the minority. Furthermore, many assume an interaction between vehicle and infrastructure sensors. Given the limitations of in-vehicle sensor deployment, more research should explore alternative strategies to implement sensors in roadside units. These units could communicate risks to drivers and VRUs through external visual or audible signals, providing a more feasible and immediate solution to enhancing VRU safety.

Finally, although simulation environments are evolving, they must continue progressing to generate scenarios increasingly similar to real-world conditions, allowing simulations to incorporate all relevant variables. For example, while CARLA is a leading tool for simulating traffic environments, its limitations must be addressed. These include accurately replicating the complexity of movements and interactions between humans and vehicles, including non-verbal communication. Additionally, improvements are needed to ensure climate variations affect VRUs, vehicles, and sensor signals as they do in real-world conditions.

## ACKNOWLEDGEMENTS

We gratefully acknowledge the support provided by the Brazilian agencies Fundação de Desenvolvimento da Pesquisa (Fundep, Rota 2030/Linha V, grant 29271.02.01/2022.04-00) and Conselho Nacional de Desenvolvimento Científico e Tecnológico (CNPq, grant 311867/2023-5).